\documentclass{article}
\usepackage[utf8]{inputenc}
\usepackage{amsmath}
\usepackage{graphicx}
\usepackage[affil-it]{authblk}
\usepackage[hidelinks]{hyperref}
\usepackage{color}
\newcommand{\vect}[1]{\mathbf{#1}}

\usepackage[numbers]{natbib}

\title{Streamlining Energy Transition Scenarios\\to Key Policy Decisions}
\author[1$\dagger$]{F. J. Baader}
\author[1$\dagger$]{S. Moret}
\author[2]{W. Wiesemann}
\author[3]{I. Staffell} 
\author[1]{A. Bardow\thanks{abardow@ethz.ch}}

\affil[1]{Energy \& Process Systems Engineering, Department of Mechanical and Process
Engineering, ETH Zürich, Switzerland}
\affil[2]{Centre for Environmental Policy, Imperial College London, United Kingdom}
\affil[3]{Imperial College Business School, Imperial College London, United Kingdom}
\affil[$\dagger$]{Contributed equally}

\date{\today}

\begin{document}

\maketitle

\begin{abstract}

Uncertainties surrounding the energy transition often lead modelers to present large sets of scenarios that are challenging for policymakers to interpret and act upon.  An alternative approach is to define a few qualitative storylines from stakeholder discussions, which 
can be 
affected by biases and infeasibilities.  
Leveraging decision trees, a popular machine-learning technique, we derive interpretable storylines from many quantitative scenarios and show how the key decisions in the energy transition are interlinked.
Specifically, our results demonstrate that choosing a high deployment of renewables and sector coupling makes global decarbonization scenarios robust against uncertainties in climate sensitivity and demand.  Also, the energy transition to a fossil-free Europe is primarily determined by choices on the roles of bioenergy, storage, and heat electrification.  
Our transferrable approach translates vast energy model results into a small set of critical decisions, guiding decision-makers in prioritizing the key factors that will shape the energy transition.  

\end{abstract} 

\clearpage
\section{Main}

As global surface temperatures have reached 1.1°C above pre-industrial levels, securing a liveable and sustainable future for humanity requires that all sectors transition urgently to deep and sustained greenhouse gas (GHG) emissions reductions~\cite{intergovernmental_panel_on_climate_change_ipcc_ar6_2023}. With energy accounting for more than two-thirds of global GHG emissions, a rapid \emph{energy transition} is at the heart of the solution to the climate challenge~\cite{lee_energy_2020_2}. 

However, despite the stark reality of climate change, the availability of mature technological solutions and clear evidence that immediate action is beneficial over a \emph{wait-and-see} strategy \cite{luderer_economic_2013, heuberger_impact_2018, riahi_cost_2021}, the Intergovernmental Panel on Climate Change's (IPCC) latest report warns that ``the pace and scale of what has been done so far, and current plans [to tackle climate change], are insufficient'' \cite{intergovernmental_panel_on_climate_change_ipcc_ipcc_2023}. 

This indecisiveness in policy deployment \cite{reiner_learning_2016}, resulting in the loss of vital time to deploy mitigation and adaptation measures, can at least in part be attributed to the substantial complexity and ``deep uncertainty'' \cite{marchau_decision_2019} affecting long-term energy systems planning and decision-making. 


\subsection*{Lost in complexity: Alternatives and uncertainty in quantitative energy system models}
This complexity can be unraveled by energy system optimization models (ESOMs) which assist policymakers in the definition of \emph{quantitative} and feasible energy transition pathways \cite{pfenninger_energy_2014}. In a nutshell, ESOMs are mathematical descriptions of energy systems that identify optimal investment and operation strategies to meet given cost or decarbonization targets \cite{contino_whole-energy_2020}. While, historically, ESOM studies have focused on generating a single optimal solution, in the recent years modelers have expanded their scope, following two main trends: 
On the one hand, evidence that cost-optimal solutions might not be meaningful proxies of real-world energy scenarios \cite{trutnevyte_does_2016} motivated modelers to explore the near-optimal solution space with the so-called modeling to generate alternatives (MGA) methods \cite{voll_optimum_2015, pickering_diversity_2022}. These methods, applied for example to the power sector in Europe \cite{neumann_near-optimal_2021, 
neumann_broad_2023}, to the US \cite{decarolis_modelling_2016} and to individual countries \cite{berntsen_ensuring_2017, lombardi_policy_2020}, identify a set of maximally different options within a small (often 5-20\%) distance from the global optimum.
On the other hand, large uncertainties in models' assumptions and input data \cite{mirakyan_modelling_2015, decarolis_formalizing_2017} -- such as fuel prices \cite{moret_characterization_2017}, demand, interest rates, etc. -- led to the integration of uncertainties in energy planning using sensitivity and uncertainty analysis methods \cite{mavromatidis_uncertainty_2018, yue_review_2018, moret_decision_2020}.

Although both MGA and uncertainty studies successfully move away from the fallacy of a ``single optimal solution'', they often present decision-makers with many scenarios and trade-offs: MGA studies result in hundreds of equivalent options to reach a given target \cite{pickering_diversity_2022, neumann_near-optimal_2021};  
uncertainty studies \cite{gabrielli_robust_2019,usher_global_2023} similarly generate hundreds of solutions by sampling possible realizations of the uncertainty.

In summary, while these methods provide valuable insights and increase the robustness of ESOMs outcomes, they 
do not give clear guidance in identifying the actual next steps in the energy transition but present instead hundreds of plausible scenarios.

\subsection*{Socio-economic storylines: Oversimplifying complexity through stakeholder engagement}

An alternative established approach to guide energy transition decisions relies on socio-economic ``storylines'', which are \emph{qualitative} descriptions of energy system configurations \cite{PatriciaFortes.2015}, typically outlined in a participatory process with relevant stakeholders \cite{WeimerJehle.2020}. 
Only after defining a few main storylines, quantitative calculations are performed using an energy model, following the ``story-and-simulation'' method \cite{PatriciaFortes.2015,WeimerJehle.2020}. 

This approach comes with important benefits: engagement of stakeholders allows accounting for ``soft'' aspects, which are difficult to capture in ESOMs, and the qualitative nature of storylines makes them accessible and interpretable to a broader public. Especially in larger research consortia, focusing on a small number of representative storylines helps to align research. The largest example are probably the IPCC scenarios \cite{scenarios2000ipcc,ONeill.2020}, where storylines have served as a common basis for scientific debate \cite{JiesperTristanStrandsbjergPedersen.2022,Hasani.2022}. 

Despite their several advantages, storylines present the risk of being biased toward the interests of dominant stakeholders and -- especially for smaller projects -- the process of stakeholder engagement is always somewhat opaque \cite{WeimerJehle.2020, Schweizer.2020, E.AndersEriksson.2022}. 
In addition, qualitative storylines received from stakeholder interviews may prove infeasible when quantified with energy system models \cite{EvelinaTrutnevyte.2014b}.

\subsection*{Closing the gap between quantitative models and storylines to inform the energy transition}

Effectively coupling the quantitative power of ESOMs with the broader accessibility of storylines has become of increasing interest in recent years to develop actionable energy transition policies.  In this context, the story-and-simulation literature has evolved towards iterative approaches to reconcile the possible discrepancies between socioeconomic storylines and model outcomes \cite{EvelinaTrutnevyte.2014b, mcdowall_exploring_2014, ElizabethRobertson.2017}. 
However, these approaches still require initial qualitative inputs as well as the definition of iteration procedures, and hence do not fully get away from the risks of biases and incompleteness associated with the socio-economic approach. 

Another research trend exploits the exploratory nature of ESOMs by generating a large number of model runs which are then aggregated into a small number of similar clusters \cite{prina_evaluating_2023}. 
Clustering can also be used in combination with ``scenario discovery'' \cite{bryant_thinking_2010, gerst_discovering_2013, kwakkel_exploratory_2017},  which uses simulation models as scenario generators to identify the combination of uncertainties under which particular model behavior occurs \cite{moallemi_narrative-informed_2017,  jafino_novel_2021, sahlberg_scenario_2021}. 
While linking uncertain inputs to clusters of scenarios helps to explore the solution space and understand the effect of uncertain drivers, it does not lead to interpretable storylines, which are needed to make decisions today.

In this paper, we present a method to streamline hundreds or thousands of energy system scenarios to a few interpretable storylines described by urgent policy decisions. 
Specifically, we show that training machine learning algorithms on key outputs of interest of ESOMs can quantitatively translate energy transition scenarios into a small number of storylines. 
The outputs of interest can be freely chosen by decision-makers to answer specific policy questions. 
Applications to both the global energy transition using an Integrated Assessment Model and to the European energy transition demonstrate that our method reduces the wide decision space to a small number of critical decisions that must be made \emph{here-and-now} to enable these transitions. 
Additionally, we propose a new way to visualize these choices into decision trees, effectively prioritizing decisions and associating each choice with its consequences. 
This representation unveils the most important interconnections and trade-offs between key policy decisions. 

Overall, our method fills the gap between quantitative and qualitative approaches to the energy transition, informing stakeholders with interpretable storylines backed up by quantitative and transparent energy system models. The software package to reproduce our results and apply our method to other case studies is available open-source at: \href{https://www.gitlab.ethz.ch/epse/systems-design-public/decide/}{https://www.gitlab.ethz.ch/epse/systems-design-public/decide/}.

\subsection{Results}

\subsubsection{Global decarbonization pathways}
\label{sec:global_case}
We first apply our method to the global decarbonization pathways under uncertainty studied by Panos et al. \cite{PANOS2023113642}, who recently presented the first Monte-Carlo assessment of an integrated assessment model (IAM). 
IAMs are large-scale models considering the energy system, the economy, and the earth's climate, that can be used to calculate global transition pathways meeting a given temperature limit by the end of the century. 
While such a long-term transition is subject to extreme uncertainties, policy decisions have to be taken now.
To address this problem, Panos et al. \cite{PANOS2023113642} consider the IPCC-SSP-2 storyline and devise energy transition pathways under uncertainty by defining 18 key uncertain parameters including energy system, economic, and climate drivers, which are then sampled in a Monte Carlo analysis. 
Here, we focus on the 1000 Monte Carlo scenarios calculated with the constraint that the global average temperature in 2100 does not exceed 2°C above the pre-industrial level. 

To demonstrate our method, we focus on model outcomes that can be translated into policies; specifically, 
we study the primary energy supply mix and the technology choices for the transition of the industry, heating and transport sectors in 2100. These outputs of interest vary by 69\% across the 1000 scenarios, which makes it difficult to derive clear guidance for these critical policy decisions from the Monte Carlo results alone. In particular, for primary energy, the renewable energy supply varies between 278 and 1119 EJ/year, and the primary fossil energy supply varies between 238 to 1279 EJ/year. Note that the model includes negative emissions technologies to compensate the emissions from fossil fuels.
For sector coupling in the three main non-electricity sectors, we consider the following outputs of interest: electrification of industry (varying between 16 and 71\%), electrification of residential heating (14 to 89\%), and the share of non-fossil transport, i.e., based on electricity, biofuels or hydrogen (46 to 87\%). A systematic procedure for the choice of the outputs of interest is provided in the Methods section (Section \ref{sec:ooi_selection}).

\begin{figure}    
    \includegraphics[width=0.9\textwidth]{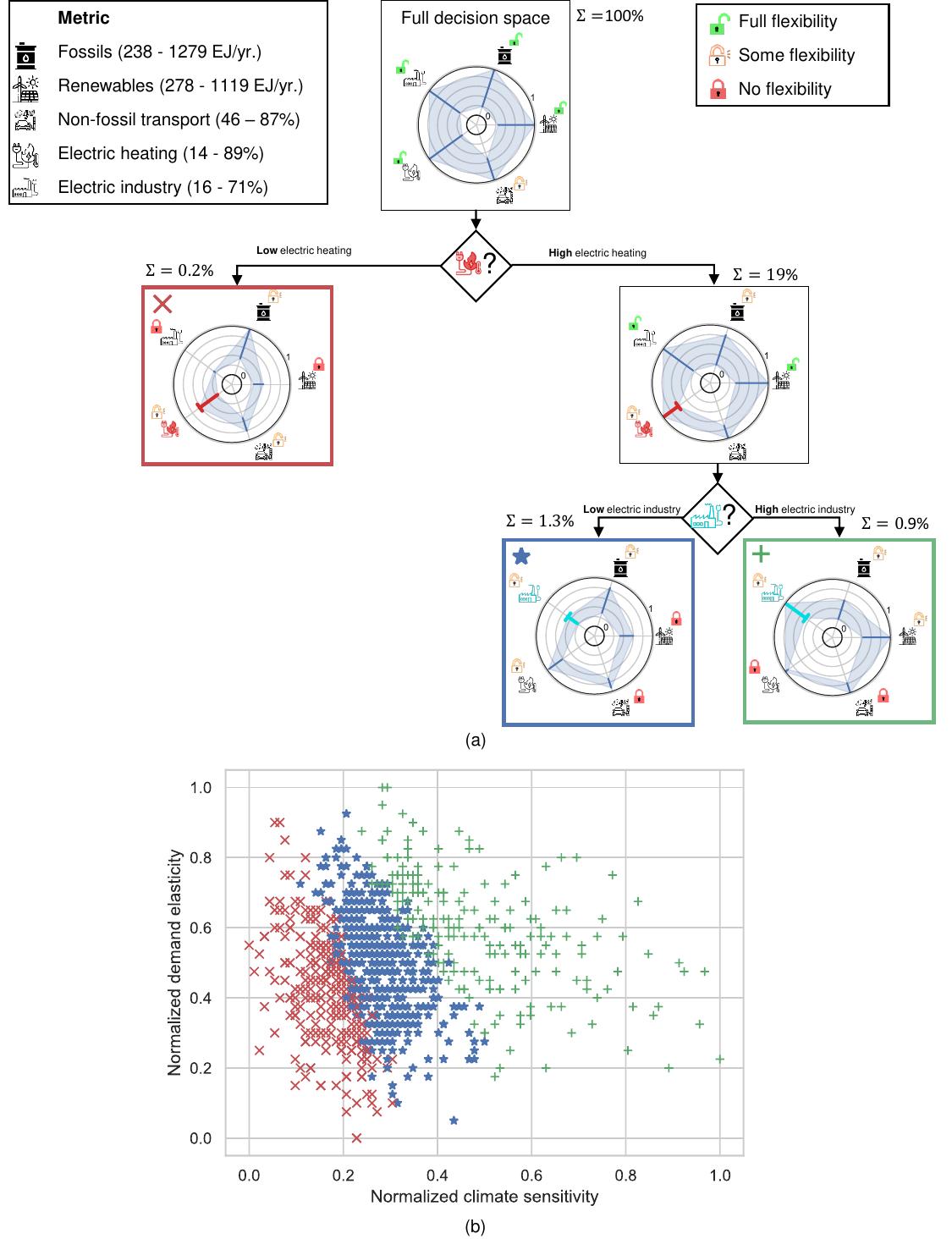}
    \caption{(a) Decision tree translating 1000 scenarios for the global energy transition in 2100 from \cite{PANOS2023113642} into two key decisions (electrification of heating and industry) along five outputs of interest. The axes of the radar plots are normalized on the range of each output of interest, while the locks indicate the level of flexibility. The size of the decision space at each node of the tree is expressed by $\Sigma$ (see Methods Equation~\ref{eq:sigma}). (b) Clustering of the 1000 scenarios along the two uncertain model inputs with the strongest effect on the resulting storyline. The coloring links the scatter plot to the three storylines of the tree in (a): Scenarios with high deployment of renewables and sector coupling are robust against uncertainties in climate sensitivity and demand.}
    \label{fig:Global_decision_tree}
\end{figure}

By applying our method, these 1000 scenarios can be summarized by two key policy decisions: (i) electrification of heating and (ii) electrification of industry (Figure~\ref{fig:Global_decision_tree} (a)). The resulting decision tree has only three leaves, corresponding to three storylines for the global energy transition, illustrating the consequences of each decision. At the root node lies the full decision space with 1000 scenarios as presented by Panos et al. \cite{PANOS2023113642}, with the radar plots and the open locks indicating the large ranges of variations in the results.

The first decision differentiates between scenarios with low and high shares of electric heating, and thereby splits the set of solutions: A low share of electric heating is found to also imply low non-fossil transport and minimal electrification of industry. 
Moreover, deployment of wind and sun is low. On the other side of the tree with a high share of electric heating, the share of non-fossil transport needs to be also high. 

As the next key decision, the tree further differentiates between solutions with a low and a high electrification of industry. 
A high electrification of industry then implies a high deployment of renewables and maximum electrification of heating. 
Overall, the extent to which the resulting storylines rely on renewable energy and sector coupling increases from left to right in Figure~\ref{fig:Global_decision_tree} (a).

The three storylines identified by the leaves of the decision tree show substantially different global energy systems and, thus, substantially different directions of global policy actions. However, given the huge uncertainty, it is crucial to understand which uncertain factors make policy actions optimal. 
Employing a scenario discovery approach \cite{gerst_discovering_2013}, we can show that 2 of the 18 uncertain factors considered by Panos et al. \cite{PANOS2023113642} are sufficient to explain how the different 
Monte-Carlo scenarios end up in one of the three storylines (Figure~\ref{fig:Global_decision_tree} (b)): (i) Climate sensitivity, expressing how strongly the global temperature reacts to greenhouse gases (i.e., for the same emissions, a higher climate sensitivity leads to a higher temperature), and (ii) Elasticity of energy service demands to their drivers, describing how strongly energy demands scale with respect to drivers such as residential floor area, industrial production index, sectoral value-added, GDP, and population. 
Figure~\ref{fig:Global_decision_tree} (b) reveals that storylines with high penetration of fossils and low sector coupling assume low climate sensitivity and low demand elasticity. With higher climate sensitivities and demand elasticities, we move towards renewables and sector coupling. 

Overall, our method breaks down the 1000 Monte Carlo results into a decision-tree with only two key decisions, translating the quantitative output of the IAM study into 
three storylines corresponding to three
actionable policies. 
Moreover, it provides the insight that betting on fossils and low sector coupling implies gambling the 2°C-target on the assumption that climate sensitivity and demand increase will be low. The storylines with high deployment of renewables and high sector coupling are robust against these uncertainties.

\subsubsection{The European energy transition}
\label{sec:European_case}
In the second case study, we consider the 441 near-optimal alternatives proposed by Pickering et al. \cite{pickering_diversity_2022} for the European sector-coupled energy system. 
All these alternatives consider an energy-self-sufficient and fossil-free Europe; thus, apart from a small fraction of nuclear power, energy is by design exclusively provided by renewable energy sources within Europe. Differently from the uncertainty-driven scenario generation in the previous case study, Pickering et al. \cite{pickering_diversity_2022} use MGA to identify near-optimal solutions, meaning that all alternatives can be chosen without making implicit bets on the realization of the underlying uncertainties. 
Still, the near-optimal alternatives show a significant maneuvering space and present an overwhelming 441 solutions to decision-makers and the general public.

Thus, for effective communication, it is essential to identify the main policy choices. As in the previous case study, we observe large variations in the potential policy decisions: (1) the share of heat electrification varies between 4 and 100\%, (2) the share of transport electrification between 53 and 100\%, (3) exploitation of bioenergy resources varies between 0 to 100\%, (4) average electricity imports per country between 4 and 69 TWh/y, and (5) installed storage capacity between 0.03 and 11 GW. 
As in the global case study, the outputs of interest address sector coupling (for the heating and transport sectors) and energy supply (bioenergy use). 
Moreover, the reliance on electricity imports relates to intensely discussed policy decisions as they show to what extent European countries produce electricity domestically or rely on an integrated electricity market. 

\begin{figure}    
    \includegraphics[width=1.0\textwidth]{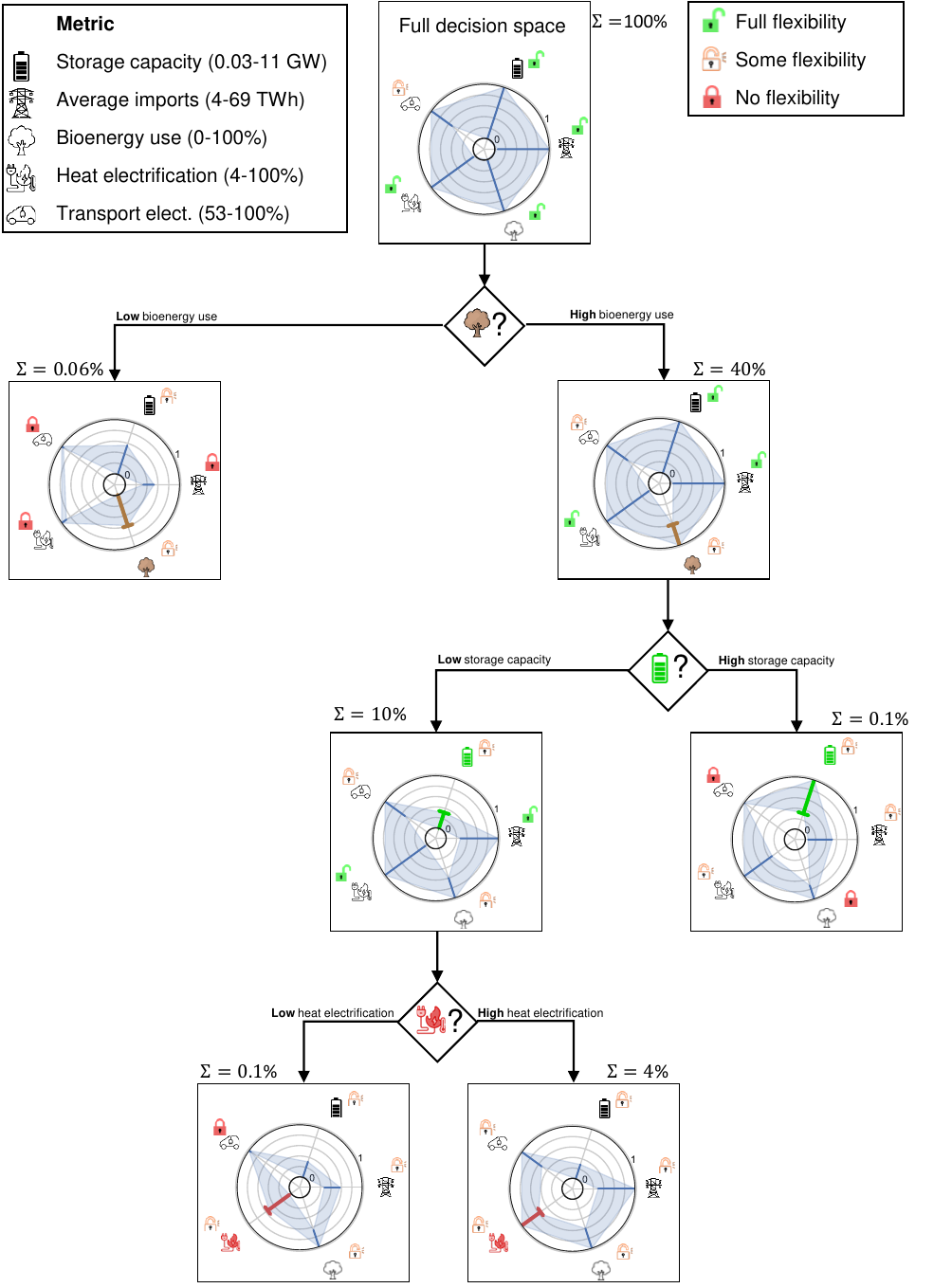}
    \caption{Decision tree breaking down the 441 alternatives for an energy self-sufficient, fossil-free European energy system presented by Pickering at al. \cite{pickering_diversity_2022} into 3 key interpretable policy choices on bioenergy use, storage, and heat electrification. For explanation of symbols, see Figure 1.}
    \label{fig:European_decision_tree}
\end{figure}

The decision tree in Figure~\ref{fig:European_decision_tree} shows that the 441 scenarios can be summarized by three decisions: (i) bioenergy use, (ii) storage, and (iii) heat electrification. 
The most critical decision is to what extent the available bioenergy should be used: Even though the entire decision space suggests freedom in all the considered outputs of interest, the choice of only using a low amount of the available bioenergy resources directly forces close-to-full electrification in both the heat and transport sectors. 

The high-bioenergy side of the tree retains the original freedom in all other outputs. 
Next, a policy decision on the installed storage capacity needs to be made, which influences national imports: 
Choosing a low storage capacity results in medium-to-high imports, and \emph{vice versa}.
Thus, with combined high bioenergy and storage uptake, countries can limit their electricity import dependency.
However, investing in storage capacity also implies 100\% transport electrification and high heat electrification to keep within the near-optimal cost range. 

Finally, if both a high use of bioenergy and a low storage capacity (thereby also medium-to-high average imports) are chosen, one must either choose low heat electrification, which implies transport electrification close to 100\%, or high heat electrification, which retains the freedom on transport electrification. 
Thus, electrification of transport can be partly avoided only if both high bioenergy use and high heat electrification are chosen. 

Overall, our decision tree clearly visualizes the two trade-offs between the five outputs of interest: 
First, bioenergy can be used to either avoid heat electrification or to avoid very high shares of transport electrification. 
Second, countries need to decide between electricity imports and storage. 

\section{Discussion}

Our method and findings have key implications for energy system modelers, socio-economic scientists, and policymakers. 

For energy system modelers, studying uncertainty or exploring the near-optimal solution space using MGA methods can lead to hundreds or thousands of scenarios on the decision-makers' table. 
Independently of how these scenarios are generated, our method can help interpret results by breaking them down into a small number of storylines. 
These storylines are organized in a clear hierarchy by decision trees splitting on the outputs of interest defined by decision-makers and thus reflecting urgent energy policy decisions. 
While clustering of solutions is not new in the energy systems domain, the novelty of training decision trees on the main outputs of interest is of pivotal importance to define actionable energy transition policies.

For socio-economic scientists, the resulting decision trees can support stakeholder discussions by clearly showing the trade-offs between storylines. For example, in the European case study, we show that it is only possible to avoid high electrification in the heating sector by exploiting most of the available bioenergy resources for energy use (Figure~\ref{fig:European_decision_tree}). We believe that these ``to avoid A you must accept B'' statements reduce the danger of unrealistic feel-good-storylines and move the discussion towards the real -- potentially unpleasant -- trade-offs that must be addressed. We do not aim to replace the socio-economic approach; on the contrary, our method starts from the definition of the key outputs of interest, which should always actively involve stakeholders following the systematic procedure detailed in the Methods (Section \ref{sec:ooi_selection}).

For policymakers, our method prioritizes the key decisions and visualizes their consequences and trade-offs. 
In the global case study, the decision tree visualizes how increasing sector coupling drives the switch from fossils to renewable energy (Figure~\ref{fig:Global_decision_tree}). 
Additionally, we show that high-renewable, high-sector coupling scenarios are a ``safe bet'' for global decarbonization with respect to climate risk and demand uncertainty. 
While we build on results by Panos et al. \cite{PANOS2023113642}, our findings are new: The decision tree identifies the critical policy decisions and connects them to the underlying uncertainties.

The European case study reveals that an energy-self-sufficient, fossil-free European energy system has some maneuvering space in the other considered policy decisions only if a high use of bioenergy is accepted. In other words, low use of bioenergy forces full electrification in all other sectors. 
Interestingly, this finding is mentioned as one exemplary finding in the original study \cite{pickering_diversity_2022}. 
The decision tree in Figure~\ref{fig:European_decision_tree} systematically shows that bioenergy utilization is not just one of the many trade-offs, but it is \emph{the} choice to be made within the proposed metrics.\\

Our study is not without several limitations. First, 
the decision trees will only visualize trade-offs between decisions that are chosen as outputs of interest.
While this implies that choosing the outputs of interest is a critical step that should always involve stakeholders, it also means that defining the outputs of interest gives the flexibility to consider specific policy questions. Additionally, the decision trees can be refined based on stakeholder feedback, helping stakeholders to quickly explore relevant trade-offs based on policy priorities. 
In the Methods (Section  \ref{sec:ooi_selection}), we present a systematic procedure to tailor the decision trees to specific policy questions that matter to stakeholders.

Second, as our method leverages machine learning techniques, the outcomes are only as good as the underlying data. 
We derived different insights from two recent peer-reviewed works, namely an uncertainty-driven global study \cite{PANOS2023113642} and a near-optimal-solution-based European study \cite{pickering_diversity_2022}. 
However, both case studies present an incomplete picture: In the first case, the near-optimal solution space is not explored. 
In the second case, we do not know how robust the solutions are with respect to uncertain input parameters, i.e., while all solutions are near-optimal for the nominal parameters, some might be more robust against uncertainties. 
However, we show that our method works in both cases, and hence it could be applied to energy system scenarios that are generated considering both uncertainty and near-optimal solutions. An additional limitation of the two case studies is that some technology options are not considered. 



Overall, our method can foster efficient communication between energy system modelers, socio-economic scientists, and policymakers, which is paramount for a swift and effective energy transition. 
Our decision trees combine quantitative information from hundreds of energy system scenarios with the intuitive interpretability of socio-economic storylines. In this way, decision-makers no longer need to choose between these two alternative approaches but can benefit from their respective added value.
As discussed, our method generates new insights by discovering structures in a dataset of energy system scenarios, irrespective of the way these scenarios are generated, and by systematically identifying key decisions and multi-dimensional trade-offs. 

\section{Methods}

In Section~\ref{sec:preliminaries}, we summarize how the scenarios used as inputs to our analyses are typically generated. In Section \ref{sec:ooi_selection} we introduce a systematic procedure to choose appropriate outputs of interest depending on the policy questions, and discuss its application to the Global and European case studies. In Section \ref{sec:creating_decision_trees} we present the method to derive the decision trees.  The software package to reproduce our results and apply the method to other case studies is available open-source at: \href{https://www.gitlab.ethz.ch/epse/systems-design-public/decide/}{https://www.gitlab.ethz.ch/epse/systems-design-public/decide/}.

\subsection{Preliminaries}
\label{sec:preliminaries}
The energy system scenarios, which are the basis for our decision trees, are typically generated from solving a mathematical optimization problem.
An optimization problem consists of an objective function~$f$, typically a cost function, which is minimized over the vector of decision variables~$\vect{x}$ (e.g., the investment and operation strategy) subject to constraints~$\vect{c}$ (e.g., resource availabilities or meeting demand): 
\begin{align}
\label{eq:opt_problem} 
& \underset{\vect{x}}{\text{min}}~~ f(\vect{x},\boldsymbol{\theta}) 
\\ \nonumber
&\text{s.t.}~~\vect{c}(\vect{x},\boldsymbol{\theta}) \leq 0
\end{align}

In Problem~\ref{eq:opt_problem}, both the objective function~$f$ and the constraint functions~$\vect{c}$ depend on parameters $\boldsymbol{\theta}$. 
These parameters $\boldsymbol{\theta}$ are input data such as prices or demands. 
Solving a linear optimization problem results in one optimal decision vector~$\vect{x}^*$. However, as discussed in the literature review, a ``single optimal solution'' might not approximate real-world energy scenarios well and hence two methods are typically used to generate alternative solutions. 
The first method considers uncertainty, i.e., accounts for the fact that the values of the parameters $\boldsymbol{\theta}$ (such as the future gas prices or energy demands) are unknown. 
Given a probability distribution for each uncertain parameter $\theta$, sampling -- often, Monte Carlo sampling -- is performed to obtain $N$ possible realizations of the uncertain parameter vector  $\boldsymbol{\theta}$. 
For instance, Panos et al. \cite{PANOS2023113642} create $N=1000$ plausible realizations for $\boldsymbol{\theta}$.
These $N$ parameter vectors $\boldsymbol{\theta}_i, i = 1, \dots, N$ lead to $N$ optimal decision vectors~$\vect{x}_i^*$, which are the inputs to our Global case study (Section~\ref{sec:global_case}).

The second method, modeling to generate alternatives (MGA), explores the space of near-optimal solutions. 
Near-optimality is defined by an acceptable cost overshoot $\varepsilon$, e.g., 10\% in the study of Pickering et al. \cite{pickering_diversity_2022}. 
Thus, all solutions $\vect{x}_i$ with an objective function 
\begin{align}
    \label{eq:near_optimal}
    f(\vect{x}_i,\boldsymbol{\theta}) \leq (1+\varepsilon)f(\vect{x}^*,\boldsymbol{\theta})
\end{align}
are accepted as near-optimal. 

Concretely, MGA studies solve a sequence of optimization problems with the objective of generating maximally different solutions $\vect{x}$. 
To this end, some key decision variables be minimized or maximized subject to the 
near-optimality constraint in Equation~\ref{eq:near_optimal}. For example, Pickering et al. \cite{pickering_diversity_2022} follow this approach to generate 441 near-optimal solutions for the European energy system, which are the input to our European case study (Section~\ref{sec:European_case}).

Overall, both uncertainty sampling and MGA generate many plausible decision vectors $\vect{x}_i$ and, thus, devise possible pathways for the energy transition. In the next subsections, we describe how our method reduces the large number of possible decision vectors $\vect{x}_i$ to an interpretable decision tree. Note that, although the studies in this work consider either uncertainty sampling or MGA, uncertainty sampling and MGA could also be applied in combination.

\subsection{Selecting outputs of interest}
\label{sec:ooi_selection}

\begin{figure}
    \includegraphics[width=1.0\textwidth]{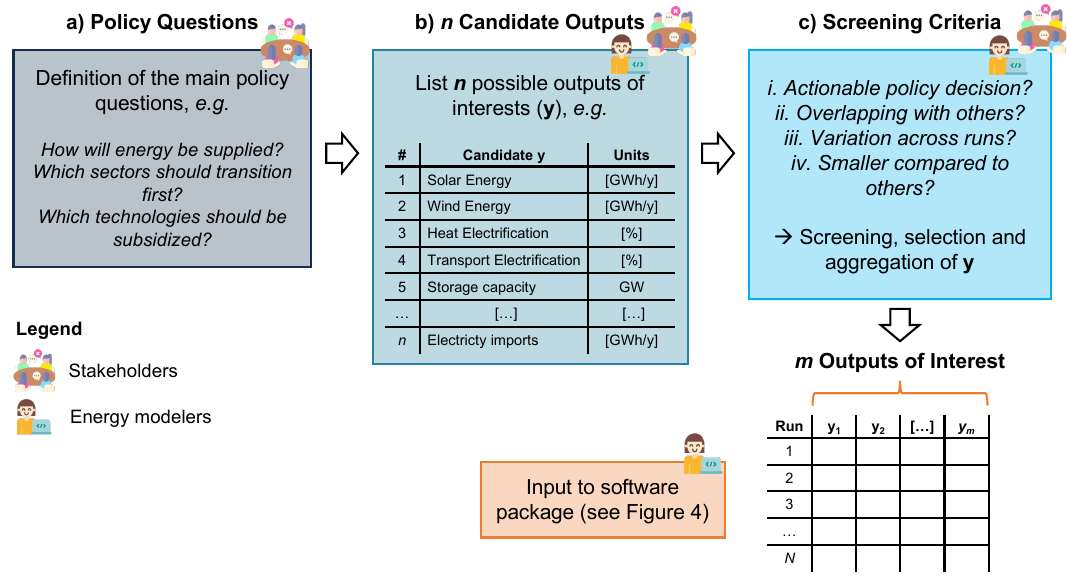}
    \caption{Systematic approach to identify outputs of interest $\vect{y}$ engaging both energy modelers and stakeholders. First, policy questions are identified by stakeholders (a). Then, stakeholders and energy modelers list $n$ (normally, 10 to 20) candidate outputs of interest which can address the identified policy questions (b). Finally, screening criteria are applied to reduce the $n$ candidates to $m$ (normally, 3 to 7) outputs of interest $\vect{y}$. The process results in a $N \times m$ matrix of scenarios which serves as input to our software package.}
    \label{fig:ooi_selection}
\end{figure}

Our method starts from the definition of $m$ outputs of interest $\vect{y}$, which can be calculated as functions of the decision variables, i.e., $\vect{y} = \vect{h}(\vect{x})$.  This step is necessary as the vector of decision variables $\vect{x}$ typically consists of thousands of outputs, while a few main policy-relevant, high-level outputs are needed to inform decision-makers. In Figure \ref{fig:ooi_selection} we present a systematic approach to help identify the outputs of interest engaging both energy modelers and stakeholders. In the Sections \ref{sec:ooi_global} and \ref{sec:ooi_europe}, we show how we applied this approach to the two case studies reported in the paper, respectively.

Step (a) concerns the definition of the policy questions, which are identified by stakeholders based on their priorities. For example, stakeholders may be interested in understanding the role played by different renewable resources in the energy transition, which technologies should be subsidized or the extent to which future energy systems will rely on imports from neighboring countries. In Step (b), energy modelers and stakeholders identify a list of $n$ possible outputs of interest that can address the identified policy questions. For example, if the policy question concerns the role played by the different renewable resources in the energy transition, the yearly electricity production by wind and solar energy are meaningful outputs of interest. In Step (c), a screening process reduces these $n$ (normally, 10 to 20) candidates to $m$ (normally, 3 to 7) outputs of interest $\vect{y}$. The first criterion is choosing actionable outputs $\vect{y}$, i.e., outputs that correspond to policy decisions: For example, if policymakers can decide to support electric mobility with subsidies or by banning fossil cars, then the share of electric mobility is a meaningful output. Second, outputs that do not vary across the $N$ are discarded. Third, outputs varying little across the runs with respect to other outputs or that are overlapping can be aggregated. Note that to allow comparison of outputs, it is important to express the outputs of interest in the same units whenever possible. In the SI Section 2.2, we additionally show how meaningful outputs of interest are connected by mass and energy balances. This systematic process finally results in $m$ outputs of interest and a $N \times m$ matrix of scenarios which serves as input to our software package.

\subsubsection{Global case study}
\label{sec:ooi_global}

For the global case study (Section \ref{sec:global_case}), we focus on two policy questions: (i) How is primary energy supplied and (ii) how are the sectors industry, heating and transport transitioned. As $n$ candidates outputs of interest, we consider the 39 outputs of interests reported in the study by Panos et al. \cite{PANOS2023113642}, to which we apply the screening criteria (Step (c) in Figure \ref{fig:ooi_selection}). First, six outputs can be excluded because they are not of interest for our policy questions (e.g., the total system cost) or because they do not correspond to actionable policy decisions (e.g., the total final energy consumption. For the first policy question, primary energy supply, we aggregate the contribution by renewables in one output of interest, i.e., (1) global primary energy supply by renewable energy, varying between 278 and 1119 EJ/year. As a second output of interest we consider the (2) global primary energy supply by fossil fuels, which also shows a significant variation across the runs (238 to 1279 EJ/year). We do not consider nuclear energy as it only varies between 6.6-7.1 EJ/y. Similarly, we exclude outputs related to the production and use of electricity and hydrogen, because they are secondary energy carriers not directly entering the primary energy balance. For the second policy question, concerning sector coupling, we focus on the transition in the three main non-electricity sectors. In this case, output metrics are expressed as relative shares in Panos et al. \cite{PANOS2023113642}, which we convert into annual energy flows. This leads to consider transport, heating and mobility as the main sectors, and to the exclusion of cooking, which represents a minor share in terms of energy flows. Aggregating the remaining outputs leads to the final three outputs of interest for sector coupling: electrification of industry (varying between 16 and 71\%), electrification of residential heating (14 to 89\%), and the share of non-fossil transport, i.e., based on electricity, biofuels or hydrogen (46 to 87\%).

\subsubsection{European case study}
\label{sec:ooi_europe}

For the European case study, Pickering et al. \cite{pickering_diversity_2022} report nine outputs in their analysis, which we consider as our $n$ candidate outputs of interest (Step (b) in Figure \ref{fig:ooi_selection}). Out of these nine outputs of interest, four are excluded because they do not correspond to actionable policy decisions. Specifically, we exclude two Gini coefficients and a Pearson correlation coefficient as these coefficients are hard to interpret for non-experts and also not directly translatable into policies.  Moreover, we do not consider the curtailment of renewables, as the maximum curtailment considered in the study is 6\%. We argue that if a curtailment of renewable energy in this order of magnitude is cost-efficient, it might not lead to public discussions. This leads to the five outputs of interest addressing the policy questions: (1) bioenergy use; (2) share of heat electrification; (3) share of transport electrification; (4) average electricity imports; (5) installed storage capacity.

\subsection{Generating decision trees}
\label{sec:creating_decision_trees}

\begin{figure}
    \includegraphics[width=1.0\textwidth]{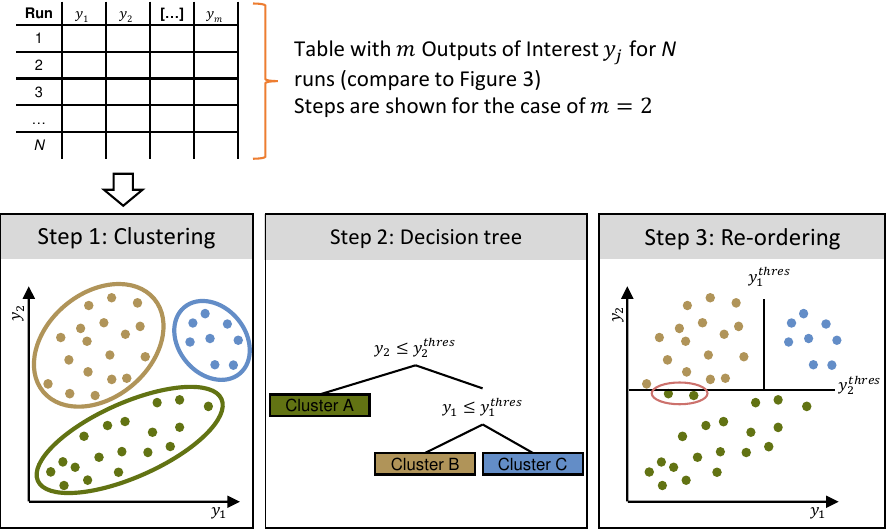}
    \caption{Conceptual example of the procedure for generating the decision trees, using only two outputs of interest: $y_1$ and $y_2$, i.e., $m=2$. In Step 1, $k$-Means clustering splits the data points into three clusters. In Step 2, a decision tree is trained to predict the cluster based on the outputs of interest $\vect{y}$. In Step 3, the points are re-ordered according to the splits learned by the tree such that a small number of points indicated with a red circle are re-assigned to a different cluster. The re-assignment is necessary as decision trees always split on a single variable, thus giving orthogonal splits.}   
    \label{fig:steps}
\end{figure}

After selecting the outputs of interest $\vect{y}$, decision trees are generated in three steps: (1) clustering, (2) training a decision tree, and (3) re-ordering the data points (Figure~\ref{fig:steps}). Note that, in the following, we represent energy system scenarios purely by their outputs of interest $\vect{y}$. 
That is, the $N$ decision vectors $\vect{x}_i$ ($i = 1, \dots, N$) generated by uncertainty sampling and/or MGA are converted to $N$ output of interest vectors $\vect{y}_i$ that define the corresponding energy system scenarios.

In Step 1, we group the $N$ different solutions to $k$ clusters where $k \ll N$. We describe how to choose $k$ in Section \ref{sec:clustering}. 
Clustering is carried out using the  $k$-Means algorithm \cite{hastie2009elements}, which has shown the best performance in the scenario discovery study by Jafino and Kwakkel \cite{jafino_novel_2021} and was used in the recent MGA study by Prina et al.~\cite{PRINA2023100100}. 
More specifically, we use the Python package \texttt{sklearn.cluster.KMeans} with default settings. 
The $k$-Means algorithm groups a set of data points, here the $N$ solutions represented by the output vectors of interest $\vect{y}_i$, to $k$ clusters, where $k$ is pre-defined. 
The rationale is to group similar points by minimizing a distance measure. 
With the standard settings, the distance measure is defined as the sum of the squared distances of each point from the cluster center. 
Consequently, the distance measure $d$ to be minimized is
\begin{align}
    \label{eq:distance_measure}
    d = \underset{\text{clusters}~c}{\sum}~~ \underset{\text{points}~i}{\sum}~~(y_{c,i} - y_c^\text{center})^2,
\end{align}
where $y_c^\text{center}$ is the center point of a cluster $c$. 
The result of the clustering is that every energy system scenario $\vect{y}_i$ is assigned to a cluster $c = 1,...,k$.

In Step 2, we interpret the clusters and identify the key policy decisions by training a decision tree \cite{hastie2009elements}. 
Decision trees are a supervised machine learning method and hence are trained on labeled data. 
In our case, the labeled data are the $N$ energy system scenarios $\vect{y}_i$, and the labels are the cluster numbers $c_i$ assigned to the data points by the previous clustering step. 
The decision tree learns to predict the label $c_i$ given the energy system scenario $\vect{y}_i$. 
For this prediction, the tree applies binary splits to the dataset, i.e., one dimension $y_k$ and a threshold  $y_k^\text{thres}$ are chosen, and the data points are split into points with $y_k \leq y_k^\text{thres}$ and points with $y_k > y_k^\text{thres}$. 
At every node of the tree, the split is chosen that minimizes the impurity of the data points. 
The impurity defines how well points within a branch of the tree are ordered, as set out mathematically in \cite{hastie2009elements}: If a branch of the tree only consists of points that belong to a single cluster, the impurity is $0$ as this branch of the tree is perfectly ordered. 
On the contrary, if the number of points belonging to a cluster is the same for all clusters, the impurity is at its maximum value. 
We limit the number of leaves of the tree to the number of clusters $k$. 
In our numerical experiments, a tree with a number of leaves equal to the number of clusters has high prediction accuracy, i.e., in both case studies, 99\% of the data points are correctly assigned to their clusters. 
To implement the decision trees, we use the python package \texttt{sklearn.tree.DecisionTreeClassifier} with default settings.

Finally, in Step 3, we re-order the energy system scenarios $\vect{y}_i$ to $k$ clusters according to the rules of the decision tree: As illustrated in Figure~\ref{fig:steps}, the few points that are predicted wrongly by the decision tree are re-assigned to the other clusters accordingly. 
With this re-assignment, the clustering becomes interpretable: Before the re-ordering, a cluster is simply a set of $m$-dimensional data points $\vect{y}_i$, where $m$ is the number of outputs of interest. 
After the re-ordering, a cluster is defined by the splits of the tree that lead to this cluster, e.g., Cluster B in Figure~\ref{fig:steps} is defined by the two conditions $y_2 \geq y_2^\text{thres}$ and $y_1 \leq y_1^\text{thres}$. 
In our numerical experiments, this re-ordering of points does not significantly affect the quality of the clustering. As shown in the SI, 
the distance measure $d$ (Equation~\ref{eq:distance_measure}) only increases by 0.1\% (Global case study) and 1\% (European case study).
If the decrease in clustering performance due to the combined clustering plus decision tree approach due to the reordering is too high for other case studies, interpretable clustering via optimal trees could be used \cite{bertsimas2021interpretable}, which combines clustering and decision tree into a mixed-integer optimization problem. 
As these optimal trees can have a high computational cost, we use a pragmatic re-ordering approach, which in our experiments shows high accuracy with low computational time.

\subsubsection{Choosing the number of clusters $k$}
\label{sec:clustering}

Finally, we discuss the choice of the number of clusters $k$. 
To identify a reasonable value for $k$, we use the elbow method \cite{ketchen1996application}, which plots the distance measure $d$ vs.~the number of clusters $k$. 
Typically, the distance measure decreases rapidly for small numbers of clusters. 
For a higher number of clusters, the distance measure only decreases slowly. 
Ideally, a so-called elbow point in between should be chosen as $k$, located where the distance measure is already small, and adding additional clusters does not substantially further reduce the distance measure. 

In this work, the elbow serves as an upper bound. 
Still, it can be reasonable to choose a smaller number of clusters to ease communication. 
As the human brain has capacity limits \cite{Miller.1956,Farrington.2011}, it is advised for effective communication not to present more than 5 alternatives \cite{Doumont.2002,Cowan.2010}. 
Moreover, the number of storylines presented in socio-economic research is typically between 2 and 5 \cite{WeimerJehle.2020}. 
Thus, it would be desirable from a communication point of view to have 5 or fewer clusters. 

As a third criterion for choosing the number of clusters, we analyze the splits of the decision tree: If the tree splits on the same output several times, the clusters can become less meaningful. 
For example, in the European case study, we choose $k=4$ which is below the elbow (the elbow curve is reported in the SI). 
With these 4 clusters, we observe 3 splits in the outputs: bioenergy use, storage capacity, and heat electrification. 
With 5 clusters, an additional split on the bioenergy use appears that does not add substantial insights, as further discussed in the SI. 
Thus, once the decision tree starts to repetitively split on the same outputs of interest, it can be reasonable to reduce the number of clusters even if this number is below the elbow.

Finally, we visualize the resulting clusters and decision tree by showing the splits and representing every branch and leaf of the tree with a radar plot showing the space covered with respect to the outputs of interest $\vect{y}$, as visualized in Figure~\ref{fig:Global_decision_tree} and \ref{fig:European_decision_tree}. 
We calculate the fraction of the original decision space covered at each branch and leaf of the tree $\Sigma$ as:
\begin{align}
\label{eq:sigma}
\Sigma = \frac{\prod_{j=1}^{m}(y_j^\text{max} - y_j^\text{min}) }{\prod_{j=1}^{m}(y_j^\text{max,initial} - y_j^\text{min,initial})},
\end{align}
where $(y_j^\text{max} - y_j^\text{min})$ is the range covered by the output of interest $j$ at one node in the tree and $(y_j^\text{max,initial} - y_j^\text{min,initial})$ is the initial range at the root of the tree. 
The number of outputs of interest is denoted as $m$.  

\section*{Acknowledgments}
F.B. and A.B. acknowledge funding by the Swiss Federal Office of Energy’s SWEET program as part of the project PATHFNDR. S.M. acknowledges support from the Swiss National Science Foundation under Grant No. PZ00P2\_202117. FB thanks Edith Baader for valuable discussions on decision trees.

\bibliographystyle{naturemag}
\bibliography{bibliography/biblio_SM,bibliography/biblio_FB}

\end{document}


\maketitle

\section{Supplementary Note 1: Global case study}
In this section, we provide additional information regarding the global case study. Specifically, we discuss the number of clusters (Subsection~\ref{sec:global_choose_k}), and the uncertainty analysis (Subsection~\ref{sec:global_uncertainty}). Finally, in Subsection \ref{sec:global_other_3000}, we extend the analysis in the main paper to the full dataset in \cite{PANOS2023113642}.

\subsection{Choosing the number of clusters $k$}
\label{sec:global_choose_k}

\begin{figure}[h]
    \includegraphics[width=1.0\textwidth]{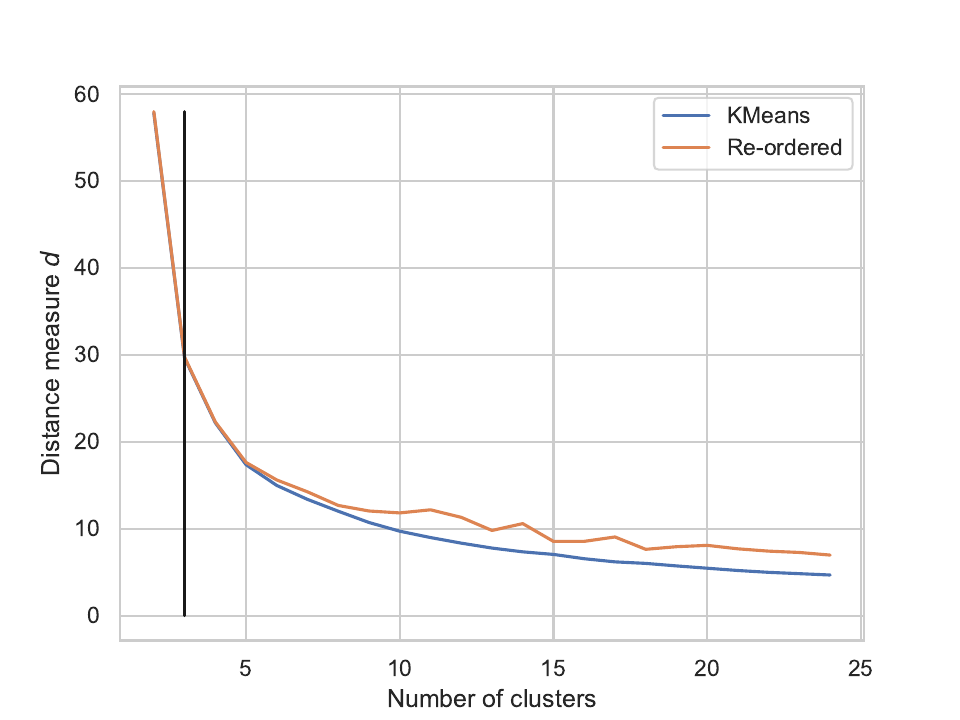}
    \caption{Elbow curve for the global case study showing the distance measure $d$ as defined in Equation~(3) of the main paper. The elbow curve is shown for both the initial $k$-Means clustering as well as for the re-ordered clusters (see  Figure 4 in the main paper). For $k=3$, the reordering of points results in an increase in the distance measure $d$ from 29.64 to 29.73 (0.3\% increase).}
    \label{fig:Global_elbow}
\end{figure}

\begin{figure}
    \includegraphics[width=1.0\textwidth]{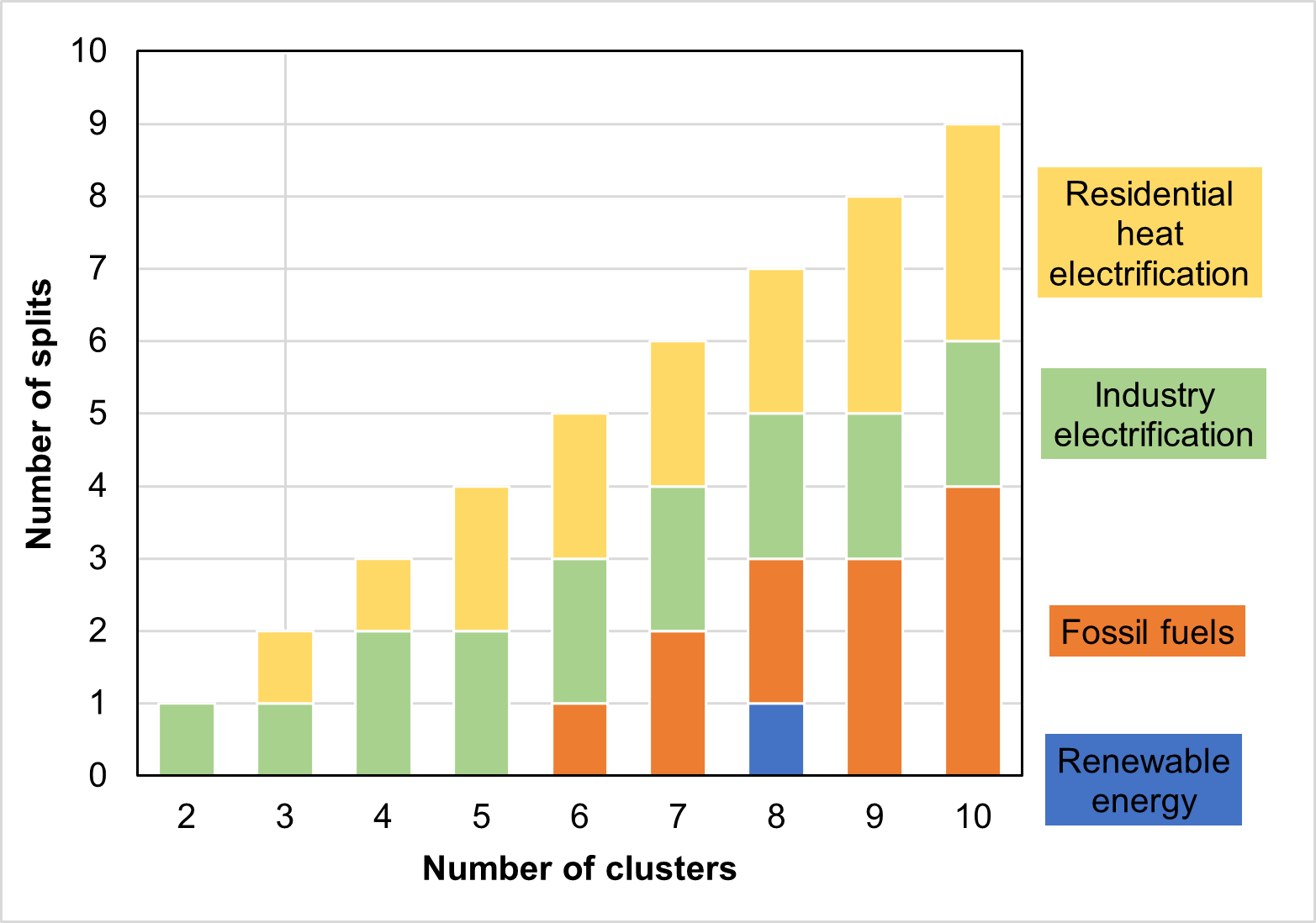}
    \caption{Number of splits depending on the number of clusters $k$ as performed by the classification tree on the outputs of interest. For up to ten clusters, the tree never splits on the share of non-fossil transport.}
    \label{fig:Global_splits}
\end{figure}

\begin{figure}    
    \includegraphics[width=1.0\textwidth]{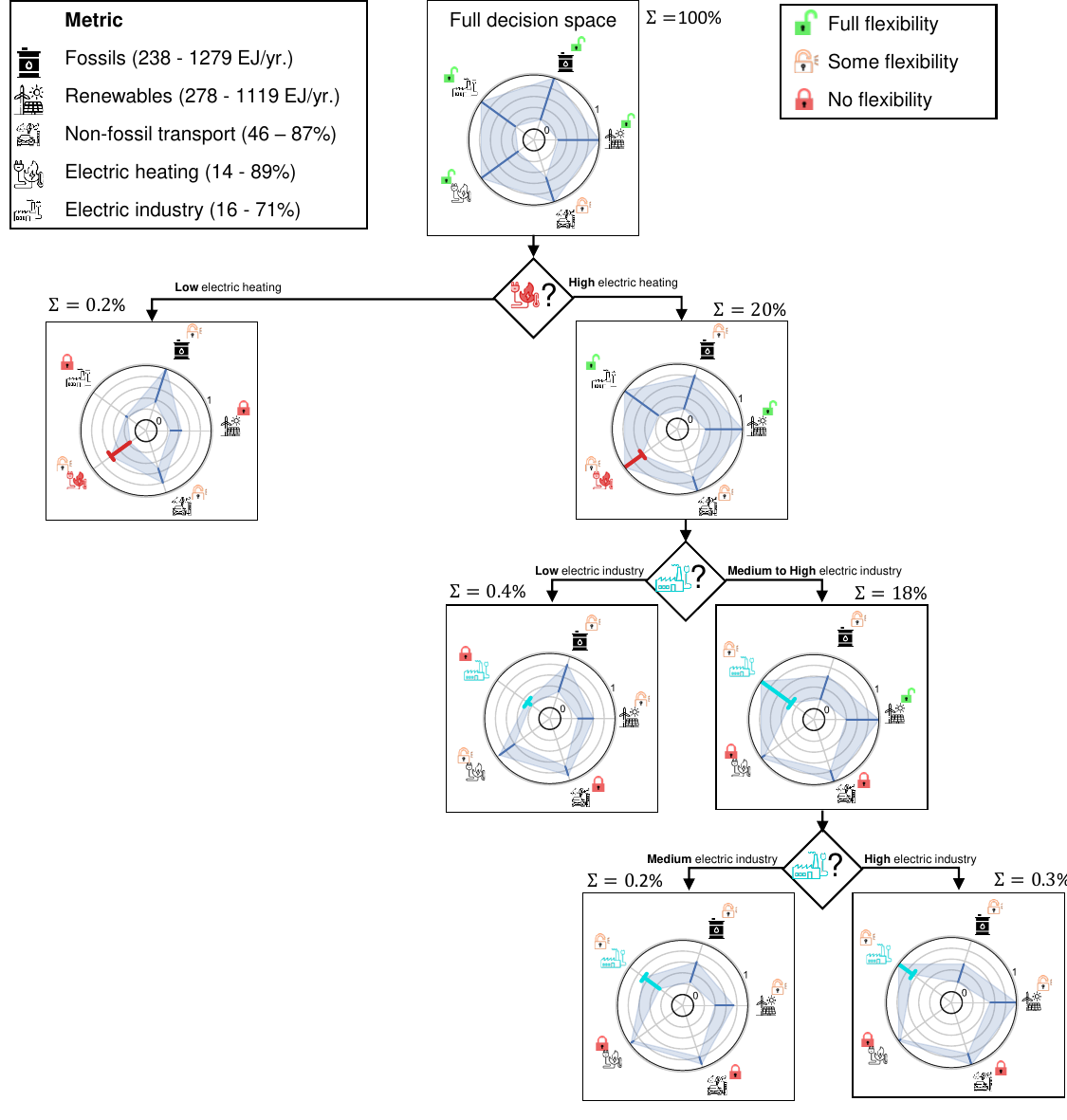}
    \caption{Decision tree for the global case study with a fourth cluster (compare to Figure 1 in the main paper).}
    \label{fig:Global_decision_tree}
\end{figure}

In this case study, we choose 3 clusters as the elbow curve shows a clear change in slope at $k=3$ (\ref{fig:Global_elbow}). Additionally, we observe that by setting $k>3$ clusters, the decision tree starts to split on the same output of interest multiple times (\ref{fig:Global_splits}). 
Specifically, with 4 clusters, the tree differentiates between low, medium, and high industrial electrification instead of only high and low. 
In our view, this additional differentiation increases the total number of clusters and hence makes communication more challenging without adding much high-level insight. 
For reference, the decision tree with a fourth cluster is shown in \ref{fig:Global_decision_tree}.

\subsection{Analyzing uncertainty by scenario discovery}
\label{sec:global_uncertainty}

To analyze which of the 18 uncertain parameters considered by Panos et al.~\cite{PANOS2023113642} are the main drivers leading to the 3 clusters we identified, we use the scenario discovery method proposed by Gerst et al.~\cite{gerst_discovering_2013}. 
In scenario discovery, a classification tree is trained to predict the cluster $c$. 
In contrast to our approach, where we train a classification tree to predict the cluster $c$ as a function of the outputs of interest $\vect{y}$, the scenario discovery approach trains a classification tree to predict the cluster $c$ as a function of the uncertain inputs $\boldsymbol{\theta}$. 

In principle, a classification tree can be trained that perfectly predicts the data; however, this decision tree will typically have a high number of leaves, making it hard to interpret. 
Moreover, a decision tree with a high number of leaves might overfit. 
Thus, Gerst et al. \cite{gerst_discovering_2013} propose to limit the number of tree leaves as a compromise between an interpretability and a coverage score. 
The interpretability score is defined as one over the number of variables that the classification tree branches on. 
Coverage is defined as the fraction of points that the tree correctly assigns to their clusters. 
This coverage is additionally calculated in a cross-validation to avoid overfitting. 
Here, we use a 5-fold cross-validation. 

\begin{figure}
    \includegraphics[width=1.0\textwidth]{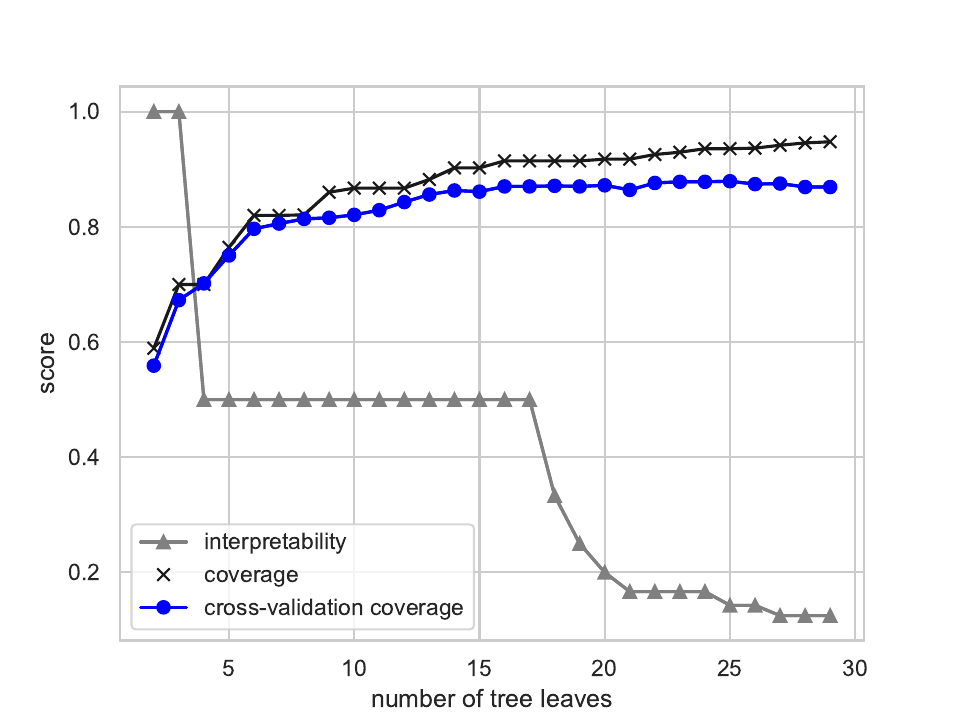}
    \caption{Interpretability score, defined as one over the number of variables that the classification tree branches on, vs. coverage score, defined as the fraction of points that are correctly assigned to their clusters by the classification tree. Moreover, the coverage of a 5-fold cross-validation analysis is shown.}
    \label{fig:Global_sensitivity_n_leaves}
\end{figure}

Analyzing interpretability vs. coverage for our case study, we observe a plateau at an interpretability of 0.5 up to 17 tree leaves (\ref{fig:Global_sensitivity_n_leaves}). 
This means that a classification tree that is allowed to make up to 16 splits of the dataset only uses 2 of the 18 uncertain parameters. 
With these 2 uncertain parameters, a cross-validation coverage of 87\% is achieved (\ref{fig:Global_sensitivity_n_leaves}), i.e., for 87\% of the scenarios, the correct cluster is predicted based on only 2 of the uncertain input parameters. These 2 parameters are climate sensitivity and demand elasticity, discussed in the main paper. 
With more than 17 tree leaves, the interpretability score drops without substantial increase in the cross-validation coverage. 
Thus, we choose a classification tree that only branches on 2 of the 18 uncertain input parameters as it is also more easily accessible than a classification tree with 17 leaf nodes. 
Additionally, having only two parameters allows for representing the decision space in a 2-D plot. The result is shown in Figure 1 (b) of the main paper,  
where we plot demand elasticity over climate sensitivity for the 1000 Monte Carlo runs. The 1000 points are colored according to the cluster they belong to in Figure 1 (a), revealing that scenarios with high deployment of renewables and sector coupling are robust against uncertainties in climate sensitivity and demand.

\subsection{Considering varying climate targets}
\label{sec:global_other_3000}
In the previous analysis and in the main paper, we considered 1000 scenarios from Panos et al.~\cite{PANOS2023113642} with the target not to exceed a 2°C temperature increase by the end of the century. 
However, Panos et al.~\cite{PANOS2023113642} study the global energy transition with three different variants:
\begin{enumerate}
    \item 2C\_SSP2: This variant sets the 2°C target at the end of the century. This is the variant used in the main paper.
    \item 2C\_SSP2\_DA30: This variant sets the 2°C target at the end of the century but delays any climate action after 2030. 
    \item 1p5c\_OS\_SSP2: This variant sets the 1.5°C target at the end of the century.
\end{enumerate}
For every variant, 1000 scenarios are generated. 
In this section, we consider all three variants simultaneously, i.e., 3000 scenarios in total, to study the effect of the different variants on the resulting storylines. Note that Panos et al.~\cite{PANOS2023113642} perform an additional 1000 runs considering the business as usual scenario without any climate action, which we do not include here.

\begin{figure}
    \includegraphics[width=1.0\textwidth]{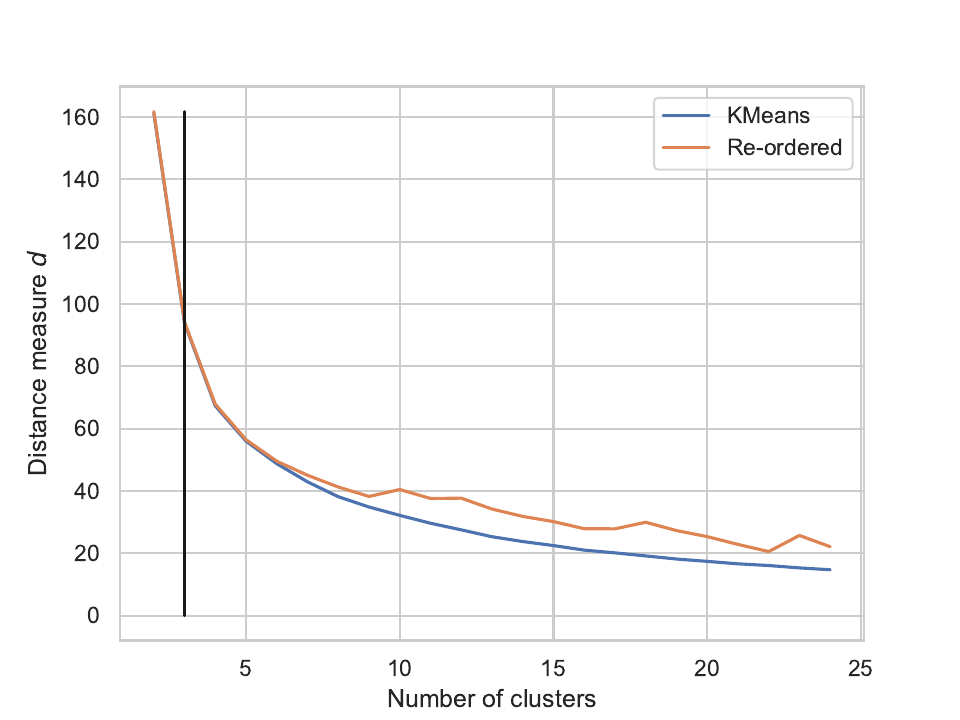}
    \caption{Elbow curve for the global case study with all three variants included showing the distance measure $d$ as defined in Equation~(3) of the main paper. The elbow curve is shown for both the initial $k$-Means clustering as well as for the re-ordered clusters (see  Figure 4 in the main paper). }
    \label{fig:Global_elbow_all_vars}
\end{figure}

First, we observe that the ranges of variations with respect to the 5 outputs of interest do not vary substantially if including all 3 variants (\ref{tab:Global_ranges}). 
Again, we use $k=3$ clusters as the elbow curve shows a nearly identical curve compared to the previous one (compare \ref{fig:Global_elbow} to \ref{fig:Global_elbow_all_vars}). 
Only the distance measure $d$ is roughly higher by a factor of 3, which is reasonable, as the number of scenarios increases from 1000 to 3000.

The resulting decision tree shows the same splits in a different order: First, the tree splits on industry electrification and only second on heat electrification (see \ref{fig:Global_decision_tree_all_vars} (a), compared to Figure 1 (a) in the main paper). 
Still, the resulting storylines are the same, as high industrial electrification implies high heat electrification (\ref{fig:Global_decision_tree_all_vars} (a)), and low heat electrification implies low industry electrification (Figure 1 (a) in the main paper). 

\begin{figure}
    \includegraphics[width=1.0\textwidth]{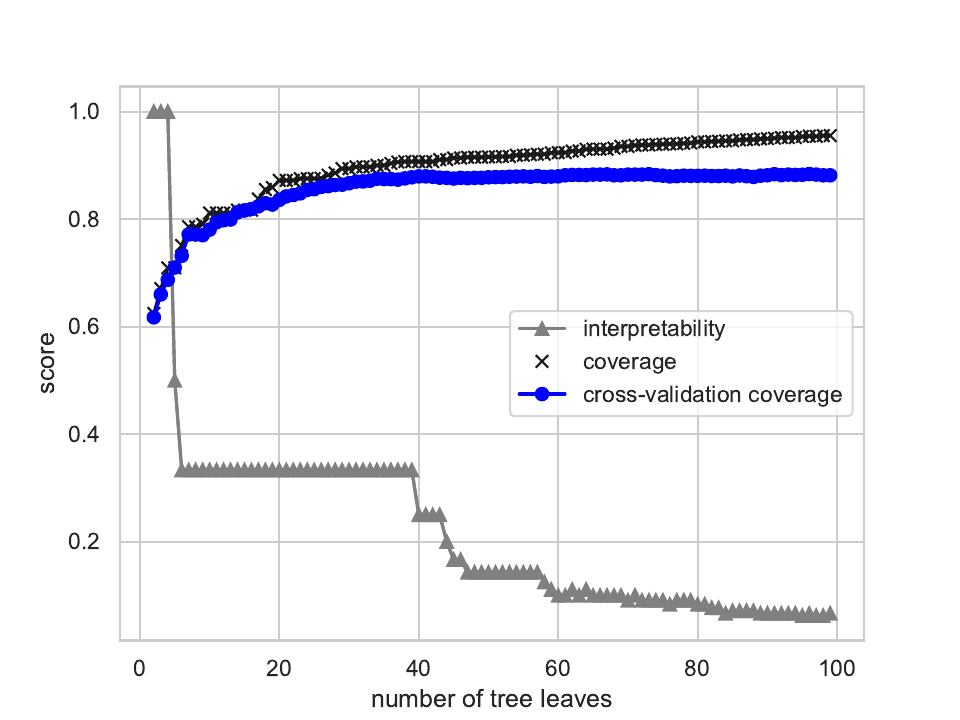}
    \caption{Scenario discovery analysis for the global case study using all three variants: Interpretability score, defined as one over the number of variables that the classification tree branches on, vs. coverage score, defined as the fraction of points that are correctly assigned to their clusters by the classification tree. Moreover, the coverage of a 5-fold cross-validation analysis is shown.}
    \label{fig:Global_sensitivity_n_leaves_all_vars}
\end{figure}

\begin{figure}    
    \includegraphics[width=1.0\textwidth]{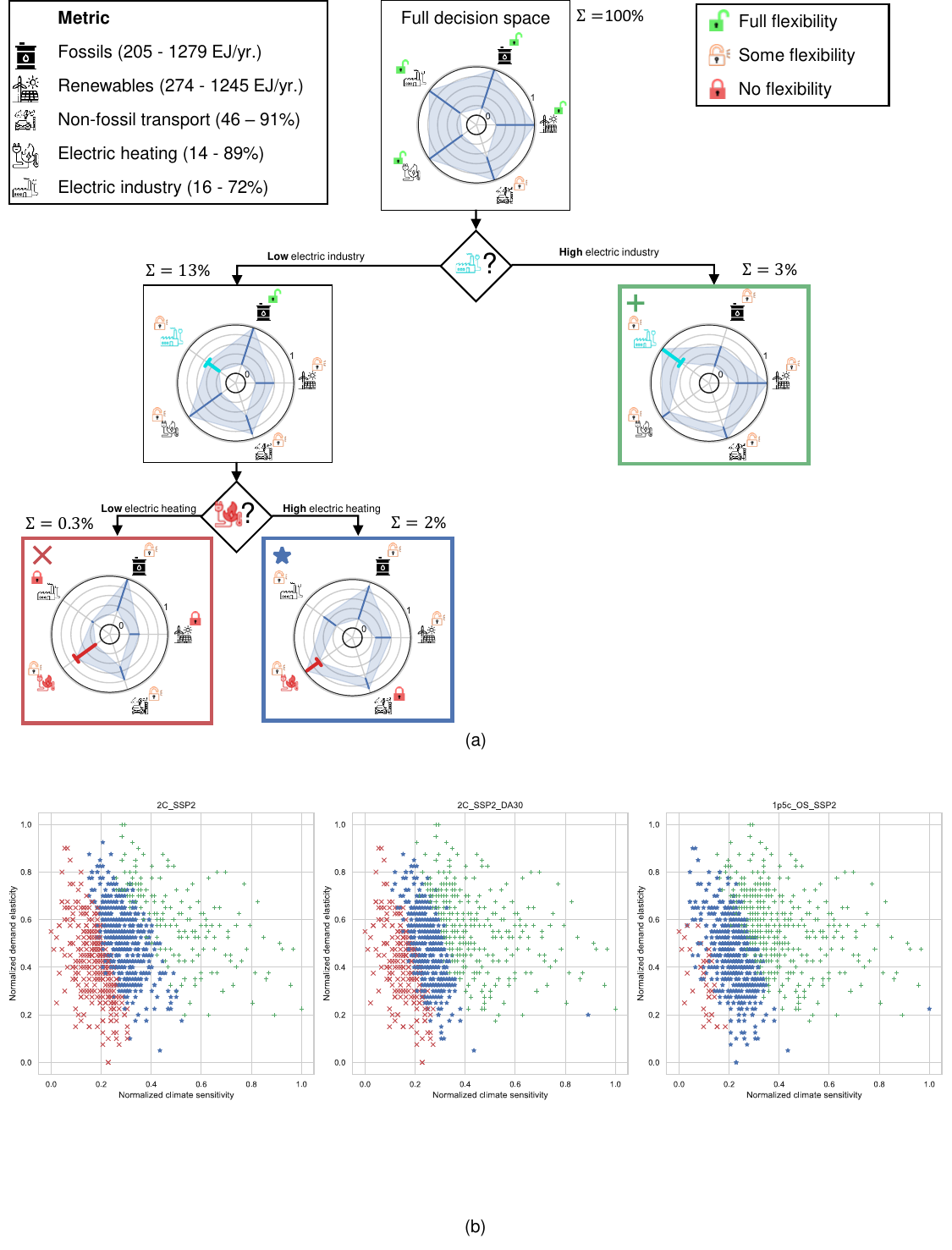}
    \caption{(a) Decision tree translating 3000 scenarios for the global energy transition in 2100 into two key decisions (electrification of heating and industry) along five outputs of interest in analogy to Figure 1 in the main paper. (b) Clustering of the 3000 scenarios along the two main uncertain drivers and the variant. The coloring links the scatter plot to the three storylines of the tree in (a): Scenarios with high deployment of renewables and sector coupling are robust against uncertainties in climate sensitivity and demand.}
    \label{fig:Global_decision_tree_all_vars}
\end{figure}

As in Subsection~\ref{sec:global_uncertainty}, we analyze the underlying uncertainties associating the Monte Carlo scenarios with the identified storylines. 
However, here, there are 19 instead of 18 uncertain inputs, as the variant constitutes a 19th uncertainty. 
A decision tree splitting on only 3 of these 19 inputs achieves a cross-validation coverage of 88\% (\ref{fig:Global_sensitivity_n_leaves_all_vars}, compare to \ref{fig:Global_sensitivity_n_leaves}). 
These 3 inputs are (i) the variant and -- as in the main paper -- (ii) climate sensitivity and (iii) demand elasticity. 
For all the three variants, we observe the same trend, i.e., betting on fossils is betting on low climate sensitivity and low demand elasticity (\ref{fig:Global_decision_tree_all_vars} (b)). 
However, the border between the three storylines shifts between the variants. 
For example, if the red storyline with a high penetration of fossils is chosen, the input space in which the 1.5°C target can be reached is much smaller (\ref{fig:Global_decision_tree_all_vars} (b)). 
Consequently, the storylines with high shares of fossils and low sector transition are even riskier if one wants to achieve the 1.5°C target. 

\clearpage

\section{Supplementary Note 2: European case study}
In this section, we provide additional information regarding the European case study. Specifically, in Subsection~\ref{sec:europe_choose_k} we discuss the number of clusters. In Subsection \ref{sec:balance_analysis}, as introduced in the main paper, we show how a quantitative analysis of energy balances can help choose the outputs of interest.
As the original energy system scenarios were created without an uncertainty sampling, the uncertainty analysis shown in Subsection~\ref{sec:global_uncertainty} cannot be applied here.

\subsection{Choosing the number of clusters $k$}
\label{sec:europe_choose_k}
\begin{figure}
    \includegraphics[width=1.0\textwidth]{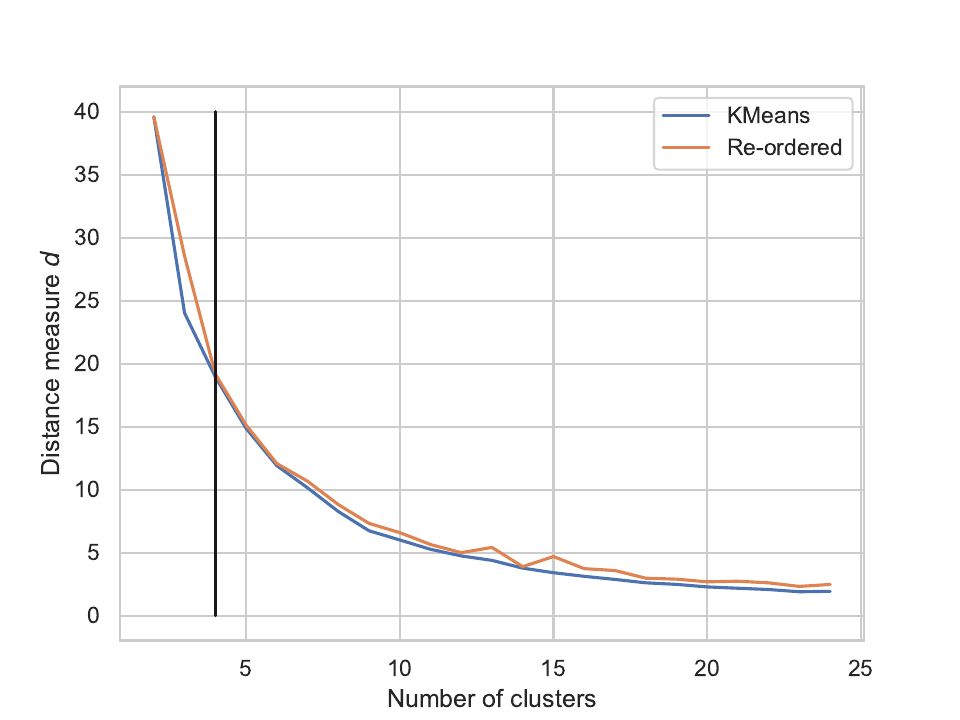}
    \caption{Elbow curve for the European case study showing the distance measure $d$ as defined in Equation~(3) of the main paper. The elbow curve is shown for both the initial $k$-Means clustering as well as for the re-ordered clusters (see  Figure 4 in the main paper). For $k=4$, the reordering of points results in a decrease in accuracy from 19.0 to 19.2.}
    \label{fig:Europen_elbow}
\end{figure}

\begin{figure}
    \includegraphics[width=1.0\textwidth]{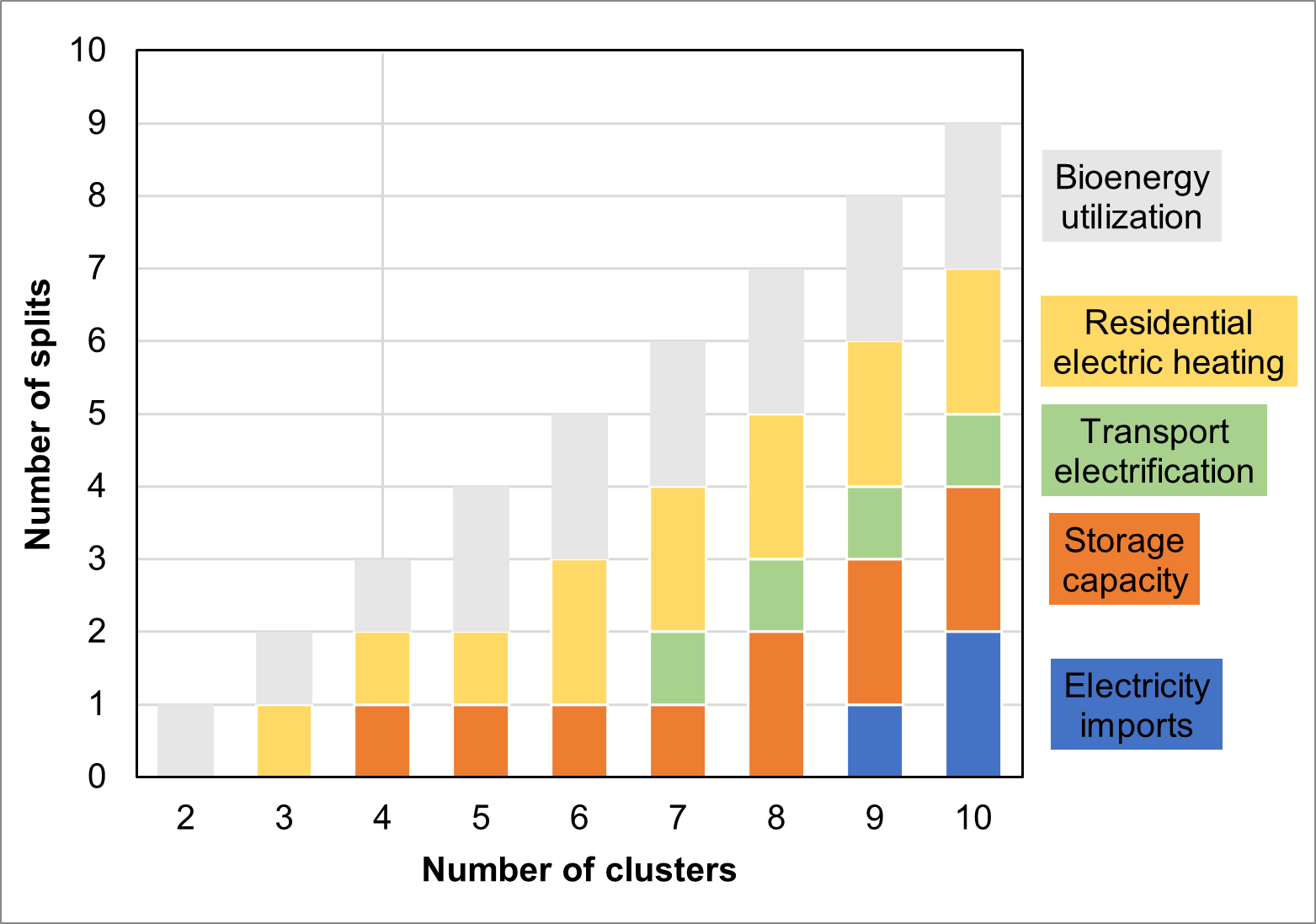}
    \caption{Number of splits depending on the number of clusters $k$ as performed by the classification tree on the outputs of interest.}
    \label{fig:Europe_splits}
\end{figure}

\begin{figure}    
    \includegraphics[width=1.0\textwidth]{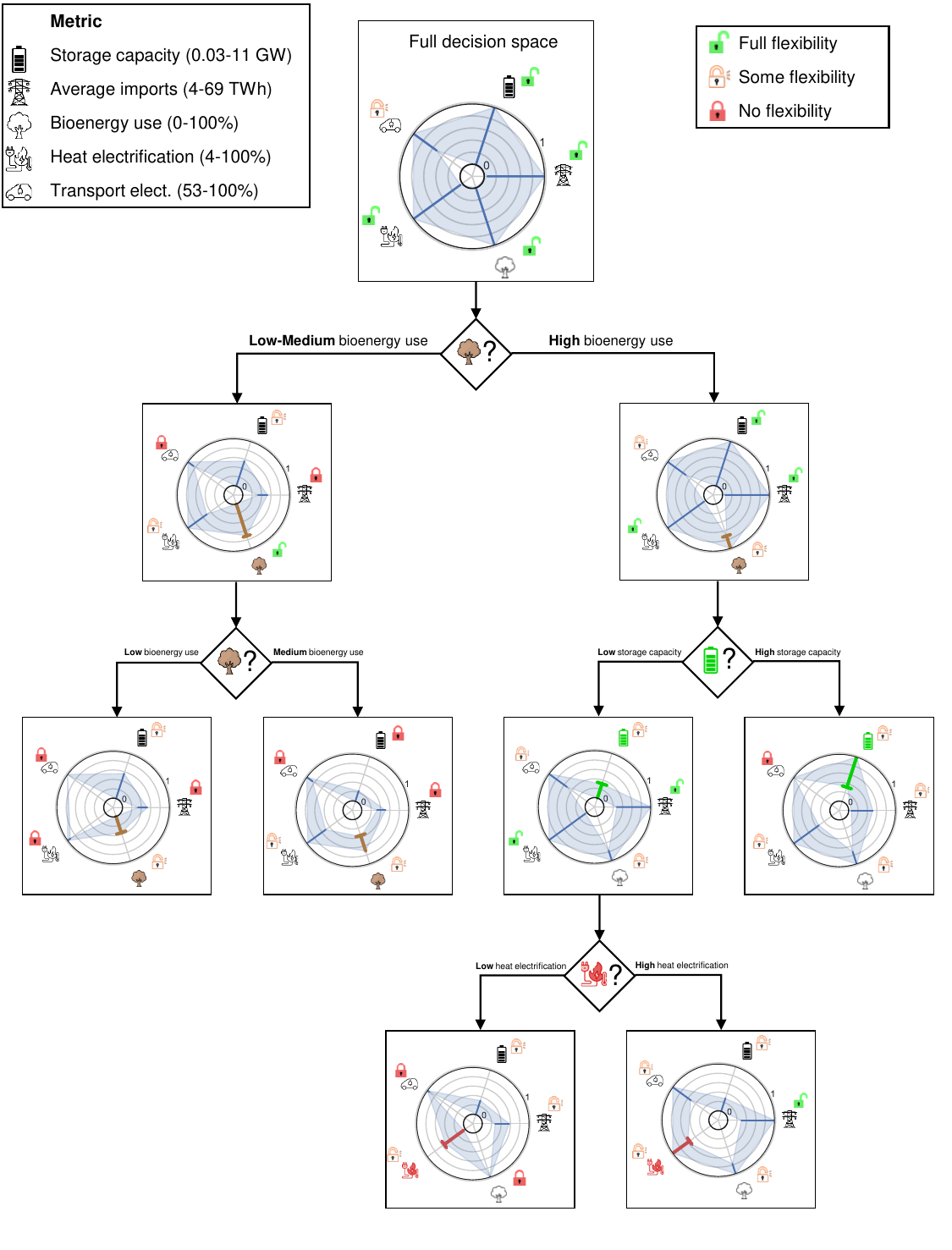}
    \caption{Decision tree for the European case study with a fifth cluster (compare to Figure 2 in the main paper).}
    \label{fig:Europe_decision_tree}
\end{figure}

In this case study, we choose 4 clusters because the elbow curve does not suggest a clear elbow-point (\ref{fig:Europen_elbow}) and with more than 4 clusters, the decision tree starts to split on the same output of interest multiple times (\ref{fig:Europe_splits}).
For example, with 5 clusters, the tree splits two times on the bioenergy use, which does not provide high-level insights. For reference, the decision tree with a fifth cluster is shown in \ref{fig:Europe_decision_tree}.

\subsection{Decision tree from a systematic screening of balances}
\label{sec:balance_analysis}

\begin{figure}    
    \includegraphics[width=1.0\textwidth]{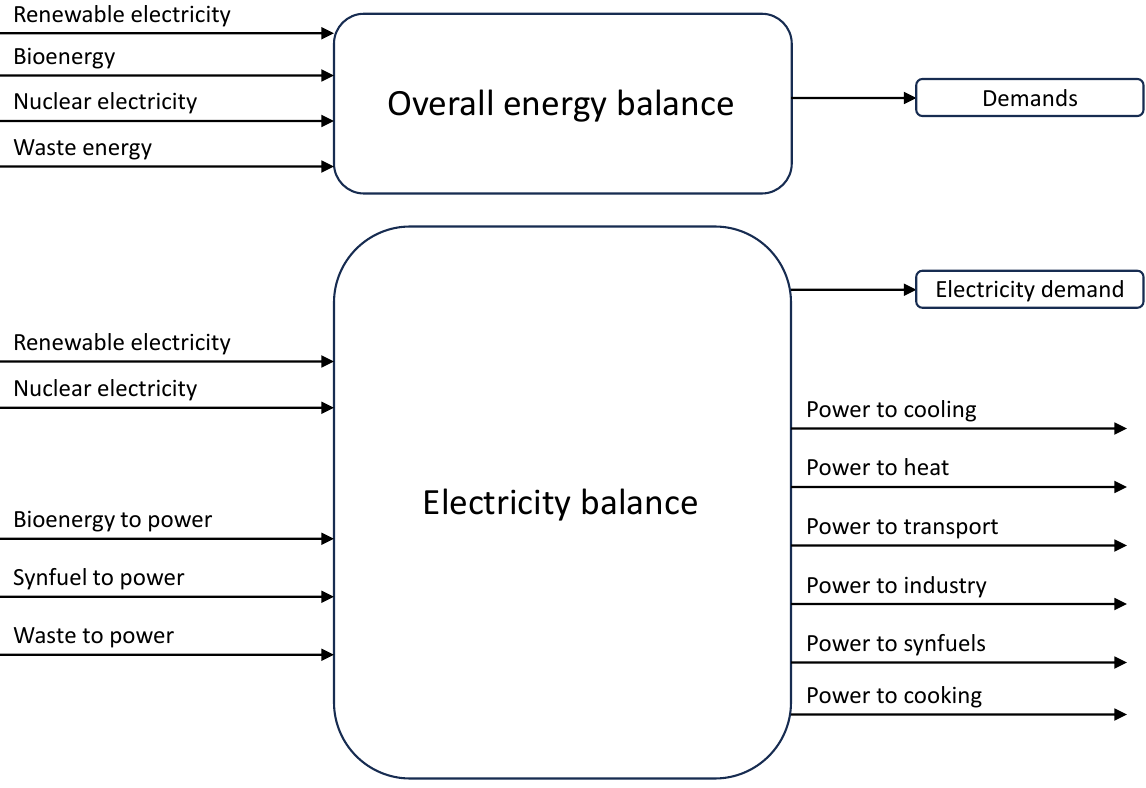}
    \caption{Considered flows in the overall energy and electricity balance for systematically choosing outputs of interest in the European case study.}
    \label{fig:Europe_flows}
\end{figure}

As discussed in the Methods section of the main paper, defining the outputs of interest is always, to some degree, subjective as it reflects the stakeholder's interest. 
Still, choosing the outputs of interest can also be done more systematically based on the observation that interesting trade-offs occur between outputs of interest that are connected by mass or energy balances. 
This is because changing one entry in a balance always causes at least a second change. Thus, outputs of interest can be derived systematically by screening the varying parts of balances. 
In this section, we show this systematic identification for the European case study, where all energy flows were published \cite{pickering_diversity_2022}. 
To this end, we consider all flows in the overall energy and electricity balance, and train a decision tree on the flows with the highest range of variation. 
We choose the overall energy balance as it shows differences in how energy is supplied, and the electricity balance as electrification is the most important measure for sector coupling. 
The two balances and the considered flows are shown in \ref{fig:Europe_flows}.

\begin{figure}    
    \includegraphics[width=1.0\textwidth]{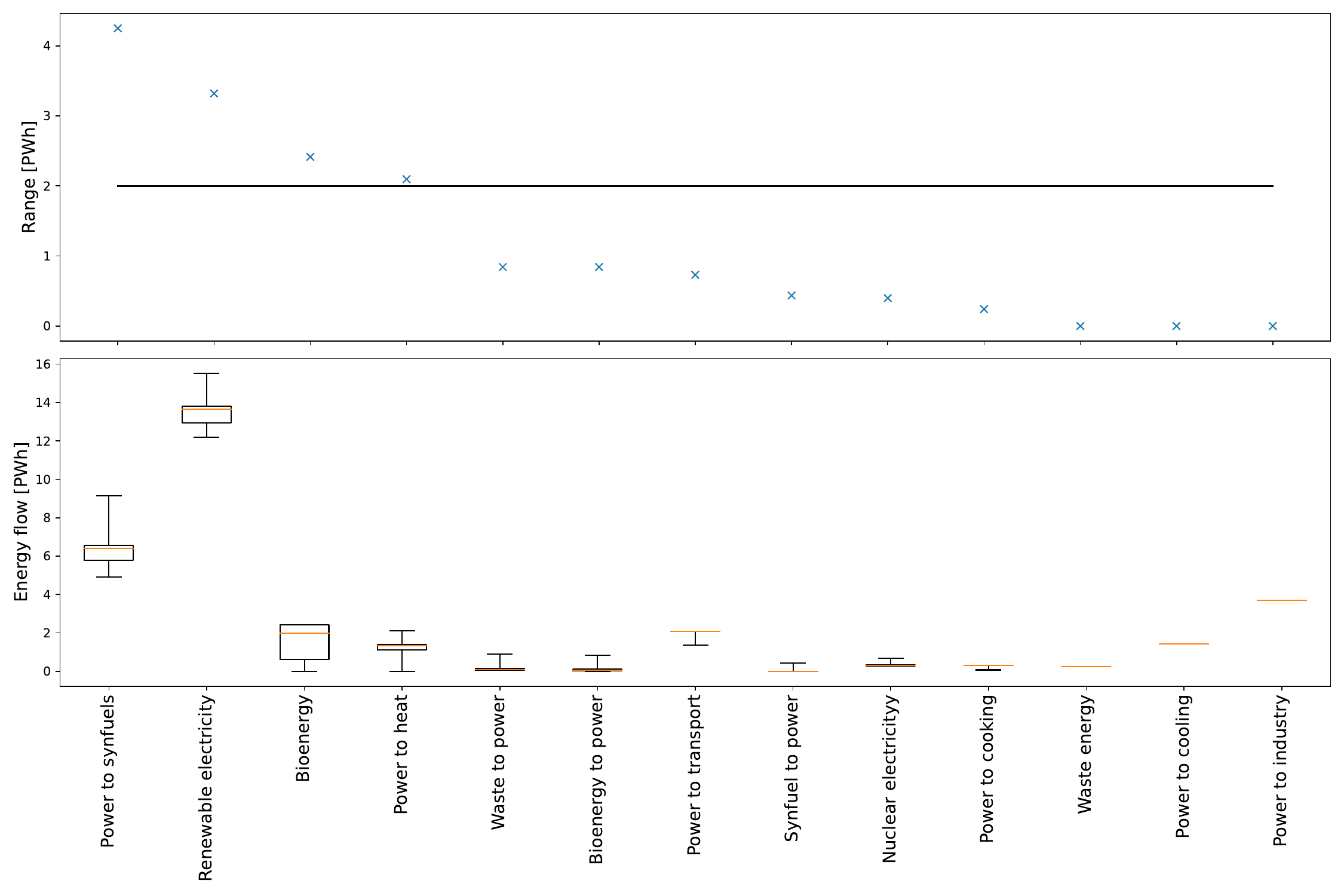}
    \caption{Ranges of variation (top) and boxplots (bottom) of the flows in the overall energy and electricity balance (compare to Figure~S9) over all 441 scenarios in the European case study. The four flows with a range above 2 PWh are chosen as outputs of interest (see the black line in the upper plot).}
    \label{fig:Europe_ranges}
\end{figure}

The four flows with the highest range over the 441 scenarios are: power to synthetic fuels, renewable electricity, bioenergy, and power to heat, all having a range greater than 2 PWh, while the next energy flow only shows a range of 0.8 PWh (\ref{fig:Europe_ranges}). 
Note that the flow ``power to synthetic fuels'' includes power used for electrolysis, hydrogen to methanol processes, hydrogen to liquid processes, and direct air capture providing carbon feedstocks for synthetic fuels.

\begin{figure}
    \includegraphics[width=1.0\textwidth]{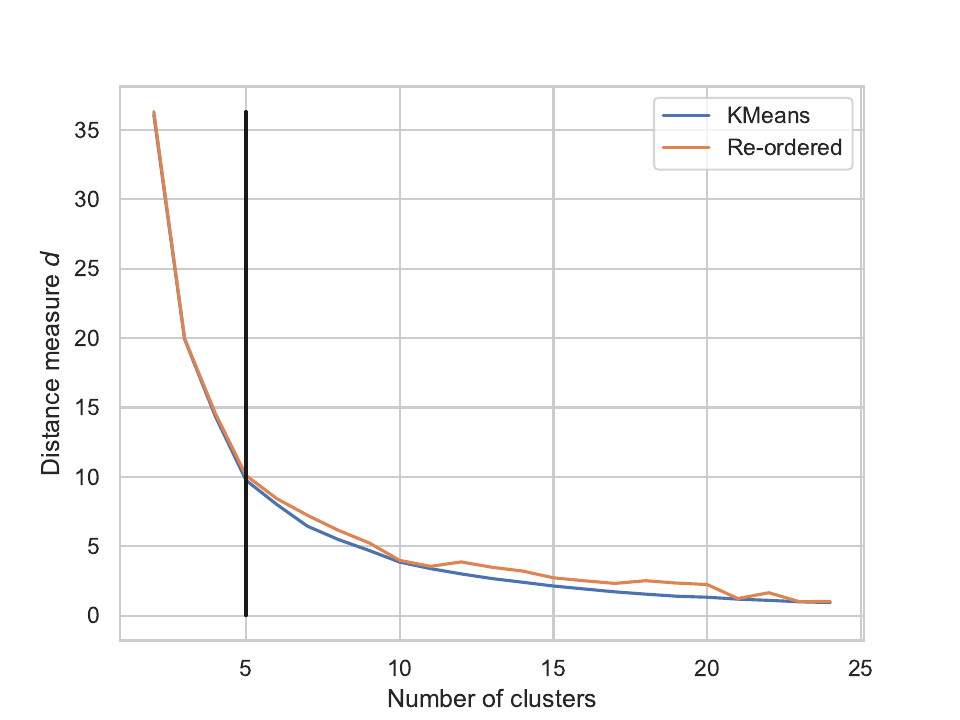}
    \caption{Elbow curve for the European case study with outputs of interest chosen from balances showing the distance measure $d$ as defined in Equation~(3) of the main paper. The elbow curve is shown for both the initial $k$-Means clustering as well as for the re-ordered clusters (see  Figure 4 in the main paper).}
    \label{fig:Europen_elbow_systematic}
\end{figure}

\begin{figure}
    \includegraphics[width=1.0\textwidth]{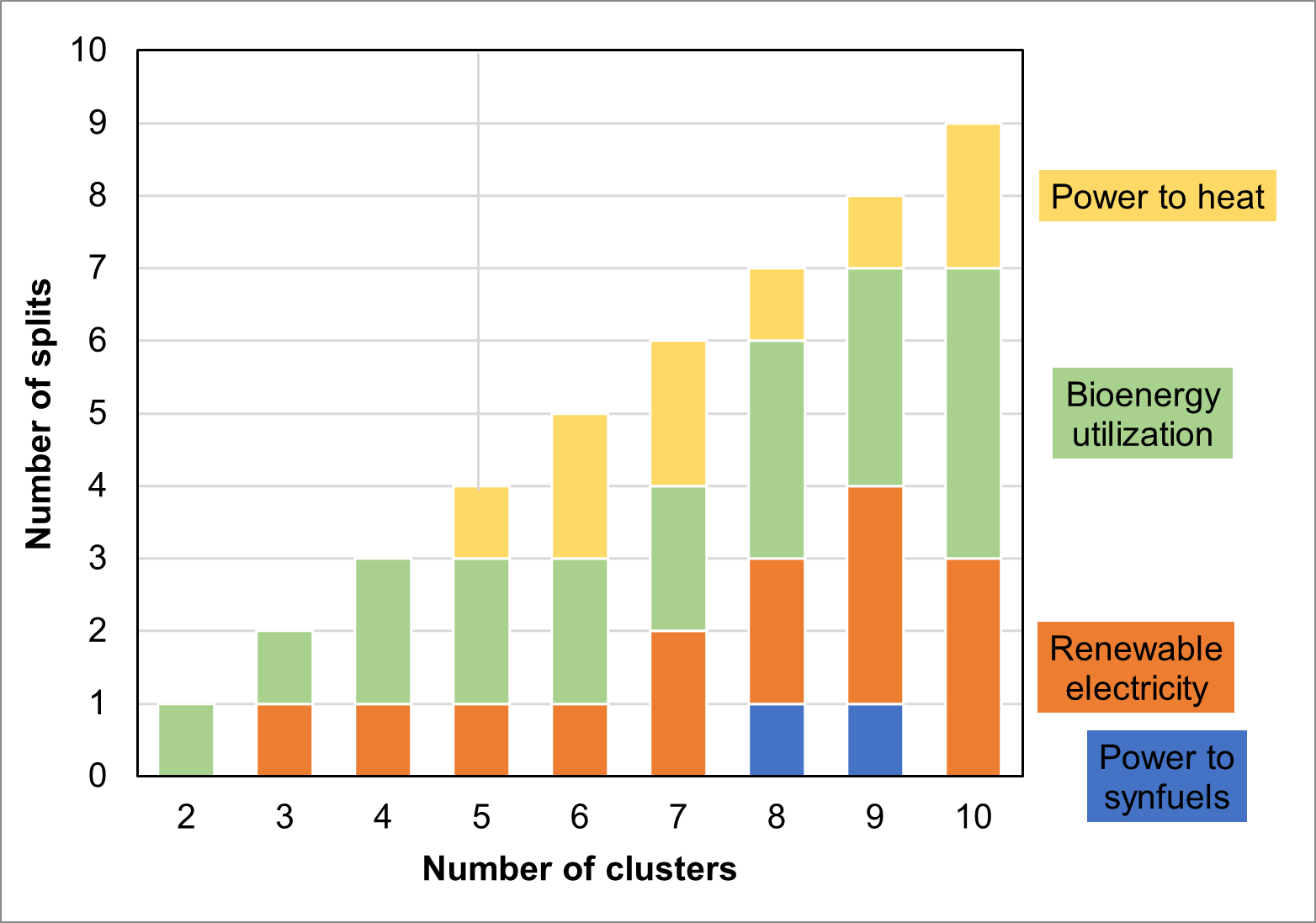}
    \caption{Number of splits depending on the number of clusters $k$ as performed by the classification tree on the outputs of interest.}
    \label{fig:Europe_splits_systematic}
\end{figure}

\begin{figure}    
    \includegraphics[width=1.0\textwidth]{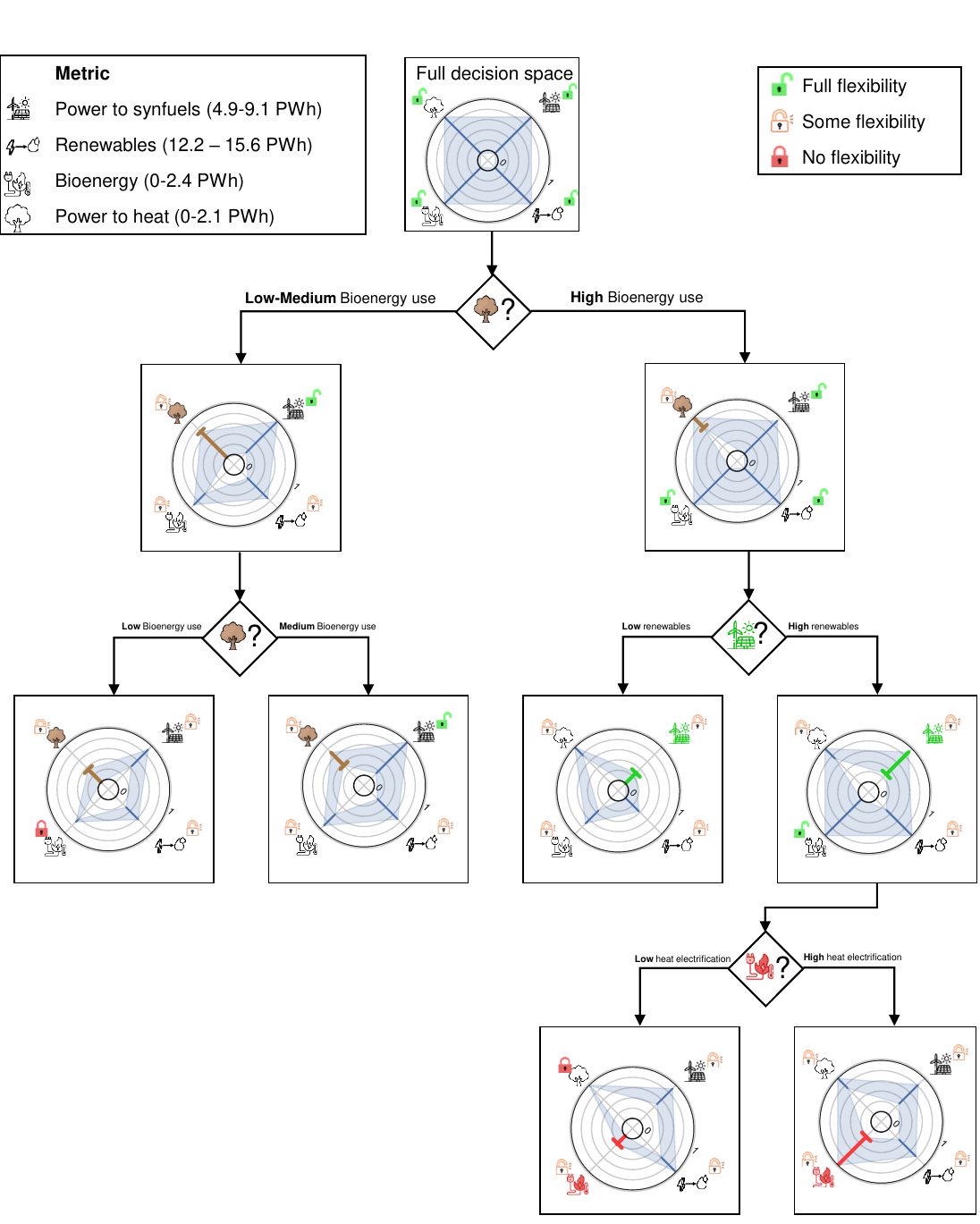}
    \caption{Decision tree for the European case study with outputs of interest chosen based on energy balances.}
    \label{fig:Europe_decision_tree_systematic}
\end{figure}

We choose $k=5$ clusters, as suggested by the elbow curve (\ref{fig:Europen_elbow_systematic}) even if this leads to 2 splits on the same output of interest, which we avoided in the previous case analysis (\ref{fig:Europe_splits_systematic}). 
The reason is that with 5 clusters the tree splits on the power-to-heat flow for the first time, which shows a trade-off that would not be present in a tree with only 3 clusters. 
The resulting decision tree is depicted in \ref{fig:Europe_decision_tree_systematic}. 
As in the main paper, the first decision is to what extent bioenergy should be used and a low bioenergy utilization shrinks the decision space in all other outputs of interest.

For high bioenergy use, a choice on the deployment of renewables must be made. 
Lower deployments of renewables lead to medium power-to-heat and low power-to-sythenthic-fuel flows. 
Higher deployment of renewables requires a further choice on power-to-heat deployment. This choice also determines the range of the power-to-sythenthic-fuel flow.

This section demonstrates that outputs of interest can be chosen systematically by screening balances. 
To use this systematic approach, stakeholders would not choose the outputs of interest directly but instead indicate their domain of interest, e.g., additional electricity demands caused by sector coupling. 
Based on these inputs, modelers can choose the balance to screen -- the electricity balance in this case -- and identify the flows with the largest range. 
Using these flows as outputs of interest results in decision trees demonstrating interesting trade-offs. 

\clearpage

\section{Supplementary Tables}
\begin{table}[hbpt]
     \centering
    \setlength{\tabcolsep}{3pt}
    \caption{Ranges of variation under the variant 2C\_SSP2, considered in the main paper, compared to the ranges in all the 3 variants in \cite{PANOS2023113642}, with respect to the outputs of interest. Changes are marked in bold.\\}
    \label{tab:Global_ranges}
    \begin{tabular}{c|c|c|c|c}
    & \multicolumn{2}{c|}{2C\_SSP2} & \multicolumn{2}{c}{all 3 variants} \\
    output of interest & min & max& min& max	 \\
    \hline
    Renewables& 	278 EJ/yr.& 	 1119 EJ/yr.& 	\textbf{274 EJ/yr.}& 	 \textbf{1245 EJ/yr.}\\
    Fossils& 	238 EJ/yr.& 	 1279 EJ/yr.& 	\textbf{205 EJ/yr.}& 	 1279 EJ/yr.\\
    Electric industry & 	16\%& 	 71\%& 	16\%& 	 \textbf{72\%}\\
    Electric heating & 	14\%& 	 89\%& 	14\%& 	 89\%\\
    Non-fossil transport & 	46\%& 	 87\%& 	46\%& 	 \textbf{91\%}\\
    \end{tabular}
\end{table}

\clearpage

\section{Supplementary Figures}

\clearpage

\clearpage

\clearpage

\clearpage

\clearpage

\clearpage

\clearpage

\clearpage

\clearpage

\clearpage

\clearpage

\clearpage

\clearpage

\clearpage

\clearpage

\bibliographystyle{naturemag}
\bibliography{bibliography/biblio_SM,bibliography/biblio_FB}